\documentclass[final,5p,times,twocolumn,authoryear]{elsarticle}

\usepackage{amssymb}
\usepackage{natbib}
\newcommand{\sortas}[1]{}
\usepackage{graphicx}
\usepackage{caption}
\usepackage{comment}
\usepackage{tabularx}

\journal{Machine Learning with Applications}

\begin{document}

\begin{frontmatter}



\title{Transforming Sentiment Analysis in the Financial Domain with ChatGPT}

\author[unipi,innov]{Georgios Fatouros}\ead{gfatouros@unipi.gr}
\author[innov]{John Soldatos}\ead{jsoldat@innov-acts.com}
\author[ote]{Kalliopi Kouroumali}\ead{kkouroumali@ote.gr}
\author[unipi]{Georgios Makridis}\ead{gmakridis@unipi.gr}
\author[unipi]{Dimosthenis Kyriazis}\ead{dimos@unipi.gr}

\affiliation[unipi]{organization={Department of Digital Systems, University of Piraeus},
            addressline={Karaoli and Dimitriou 80}, 
            city={Piraeus},
            postcode={18534}, 
            country={Greece}}

\affiliation[innov]{organization={Innov-Acts Limited},
            addressline={Kolokotroni 6}, 
            city={Nicosia},
            postcode={1101}, 
            country={Cyprus}}

\affiliation[ote]{organization={Hellenic Telecommunications Organisation S.A.},
            addressline={Kifissias Avenue 99}, 
            city={Athens},
            postcode={15124}, 
            country={Greece}}

\begin{abstract}

Financial sentiment analysis plays a crucial role in decoding market trends and guiding strategic trading decisions. Despite the deployment of advanced deep learning techniques and language models to refine sentiment analysis in finance, this study breaks new ground by investigating the potential of large language models, particularly ChatGPT 3.5, in financial sentiment analysis, with a strong emphasis on the foreign exchange market (forex). Employing a zero-shot prompting approach, we examine multiple ChatGPT prompts on a meticulously curated dataset of forex-related news headlines, measuring performance using metrics such as precision, recall, f1-score, and Mean Absolute Error (MAE) of the sentiment class. Additionally, we probe the correlation between predicted sentiment and market returns as an addition evaluation approach. ChatGPT, compared to FinBERT, a well-established sentiment analysis model for financial texts, exhibited approximately 35\% enhanced performance in sentiment classification and a 36\% higher correlation with market returns. By underlining the significance of prompt engineering, particularly in zero-shot contexts, this study spotlights ChatGPT's potential to substantially boost sentiment analysis in financial applications. By sharing the utilized dataset, our intention is to stimulate further research and advancements in the field of financial services.
\end{abstract}


\begin{highlights}
\item A pioneering effort in evaluating the capability of large language models, specifically ChatGPT, for sentiment analysis in the financial domain, focusing primarily on the foreign exchange market.

\item A curated dataset of forex-related news headlines publicly available. This open-source dataset will serve as a valuable resource for future research, stimulating further advancements in the field of financial services.

\item  An in-depth evaluation of ChatGPT's potential in financial sentiment analysis. Multiple prompts are employed for both sentiment class prediction and sentiment score estimation, considering the analysis of both single and multiple news headlines, showcasing ChatGPT's adaptability and competence in handling complex financial texts.

\item ChatGPT demonstrated considerable advancements over FinBERT, a well-regarded sentiment analysis model in financial texts. It showed an improvement of approximately 35\% in sentiment classification performance and a significant enhancement of about 36\% in correlation with market returns, underlining the extensive potential of ChatGPT in financial sentiment analysis.

\item ChatGPT exhibits significant potential for enhancing the accuracy and depth of sentiment analysis in financial services. The study underscores the importance of prompt engineering, particularly in zero-shot contexts, in optimizing performance.

\end{highlights}


\end{frontmatter}


\section{Introduction}
\label{sec:1}

\subsection{Background and Motivation}
The financial services sector has long been an early adopter of technological advancements, continually evolving to meet the demands of a rapidly changing global landscape.  From the introduction of ATMs (Autmatic Teller Machines) and electronic trading to the rise of financial technologies (Fintech), the financial sector has witnessed a transformative shift, particularly with the integration of artificial intelligence (AI) and machine learning (ML) technologies \citep{arner2015evolution}. The Fintech market has seen significant growth, demonstrating the impact of these novel technologies in finance \citep{mordor22}. Such services include credit scoring and risk assessment, fraud detection and prevention, robo-advisory services as well as chatbots for customer service and support \citep{fatouros2023deepvar, kotios2022deep}. 

Among the many facets of this transformation, sentiment analysis has emerged as a critical tool for understanding market dynamics and predicting future trends. The ability to analyze sentiments can provide valuable insights into the collective mood of the market, thus enabling more informed and strategic decision-making.

Traditionally, financial sentiment analysis has relied on manually curated lexicons and simple ML algorithms \citep{schumaker2009textual}. However, with the rapid progress in Natural Language Processing (NLP), more sophisticated techniques are now available. Deep learning-based models, such as BERT (Bidirectional Encoder Representations from Transformers) \citep{devlin2018bert}  and its financial domain-specific counterpart FinBERT \citep{liu2021finbert}, have significantly improved sentiment analysis accuracy and reliability.

Despite these advancements, the financial domain presents unique and multifaceted challenges to sentiment analysis. Financial texts, particularly news headlines, are often saturated with domain-specific terminology and nuanced sentiments, complicating the task of sentiment classification \citep{loughran2011liability}. A typical characteristic of financial text is the interweaving of multiple topics, such as different financial instruments, in the same context. Consider, for instance, the headline "CAD is one of the better places to sell the US dollar on a pullback." Conventional sentiment analysis tools, might predict a positive sentiment for this statement, which is accurate concerning the Canadian dollar but misleading for the forex pair USD/CAD. Such models mainly focus on sentiment classification, often failing to infer the subject of the text, i.e., the financial instruments related to the sentiment, which is a crucial aspect of financial analysis.

In addition to this, conventional sentiment analysis models often lack the ability to adjust their output based on specific use-case context, further limiting their broader applicability \citep{poria2016aspect}. For example, the sentiment expressed in a discussion about a new governmental regulatory policy may differ significantly depending on whether the context is an investor forum discussing potential market impacts or a consumer forum discussing changes in service fees. Typical sentiment analysis models may fail to capture these contextual nuances effectively, offering room for more context-aware models  to provide more comprehensive and accurate sentiment analysis in the financial domain \citep{poria2017review}. Thus, addressing these limitations and meeting these specific challenges head-on represents a significant area of opportunity for advancements in financial sentiment analysis.

Among the various AI tools, Generative Pre-trained Transformers (GPT) and their related technologies have demonstrated significant potential in revolutionizing multiple domains, including the financial sector \citep{george2023review}. GPT, a state-of-the-art language model developed by OpenAI, is trained on an extensive corpus of data, enabling it to understand and replicate patterns in human language with remarkable precision \citep{radford2018improving}. Since the introduction of the initial GPT model in 2018, OpenAI has consistently improved the architecture by adding new layers, implementing technical adjustments, and extending training on additional datasets, culminating in the latest and most advanced version, GPT-4 \citep{openai2023gpt4}. However, the broader impact of GPT on the public became evident towards the end of 2022 with the release of ChatGPT, a specialized version of the GPT architecture fine-tuned for conversational applications.

ChatGPT offers promising opportunities to improve existing financial applications, such as risk analysis through sentiment analysis. By harnessing the power of ChatGPT's advanced natural language understanding capabilities, financial institutions can significantly enhance the accuracy and depth of sentiment analysis. With its ability to comprehend and interpret complex language patterns, it can more effectively analyze vast amounts of unstructured data, such as news articles and headlines, to provide a comprehensive understanding of market sentiment. This improved sentiment analysis can, in turn, inform decision-making in areas such as investment strategies, risk management, and portfolio optimization, leading to better-informed decisions and potentially higher returns \citep{tetlock2007giving, chen2014wisdom}. Additionally, ChatGPT's conversational abilities can be utilized to communicate and explain complex risk analysis findings to both expert and novice users, making financial insights more accessible and actionable \citep{yue2023democratizing}.

These ChatGPT's advanced natural language understanding capabilities carry potential for a wealth of innovation in the financial domain, including financial sentiment analysis. This area, despite its significant implications, remains largely untapped, signaling an open avenue for research and development. The capacity of ChatGPT to interpret complex language patterns, a characteristic often found in unstructured financial data, presents a promising solution for enhancing the depth and accuracy of sentiment analysis in financial contexts. 

\subsection{Value Proposition}

Acknowledging the noticeable lack of research, this work explores the potential of large language models (LLMs), particularly ChatGPT 3.5, in the realm of financial sentiment analysis, with a primary focus on the foreign exchange market (forex). A zero-shot prompting approach is employed in this study, enabling us to assess ChatGPT's capabilities in deciphering financial text, eliminating the need for domain-specific fine-tuning.

To ensure a reliable evaluation, we have meticulously curated and annotated a dataset of forex-related news headlines. This dataset, made publicly available, facilitates a comprehensive evaluation of various ChatGPT prompts, thereby highlighting the model's extensive training on diverse datasets and its inherent adaptability in different sectors, including finance. By sharing this annotated dataset, we aim to contribute a valuable resource to the research community, fostering further advancements and innovation in the field of financial services and particularly in financial sentiment analysis.

The evaluation goes beyond traditional metrics, including not only precision, recall, f1-score, and Mean Absolute Error (MAE) of the sentiment class, but also the correlation between predicted sentiment and market returns \citep{hossin2015review}. This detailed evaluation extends to sentiment class prediction and sentiment score estimation for both single and multiple news headlines, demonstrating ChatGPT's agility and proficiency in managing intricate financial texts.

Moreover, by benchmarking multiple prompts, this study underscores the pivotal role of prompt engineering, especially in zero-shot contexts, to optimize performance and boost the efficacy of sentiment analysis. By shedding light on the enormous potential of ChatGPT, this manuscript aims to instigate further advancements and innovation in financial services.

The remainder of this paper is structured as follows: Section \ref{sec:2} reviews prior work on sentiment analysis in finance and discusses the role of AI tools such as ChatGPT and FinBERT in this domain. Section \ref{sec:3} details our methodology, highlighting our approach to using and evaluating ChatGPT for sentiment classification in the forex market. Section \ref{sec:4} presents our findings, which include performance metrics of sentiment classification and sentiment correlation with market returns. In Section \ref{sec:5}, we discuss the implications of our results, the potential applications of ChatGPT in financial services, as well as future research directions. The paper is concluded in Section \ref{sec:6} with a summary of our study's key contributions.

\section{Related Work}
\label{sec:2}

\subsection{Sentiment Analysis in Finance}
Sentiment analysis, or opinion mining, refers to the use of computational techniques to extract subjective information, such as opinions and sentiments, from textual data \citep{bing2012sentiment}. It has been applied to various domains, including customer reviews, social media posts, and news articles. In finance, sentiment analysis is used to gauge market sentiment, an essential aspect that could impact the price movement of financial instruments \citep{bollen2011twitter, siering2018explaining}.

The idea that sentiment or public mood can influence financial markets dates back to the era of John Maynard Keynes \citep{keynes1937general}, who described financial markets as being driven by "animal spirits" or waves of pessimism and optimism. This notion has been substantiated by numerous empirical studies, demonstrating that changes in investor sentiment could significantly affect asset prices and trading volume \citep{baker2007investor, tetlock2007giving}.

Initially, financial sentiment analysis relied on manually curated financial lexicons, where words were pre-labeled as positive, negative, or neutral \citep{loughran2011liability}. However, this approach was often too simplistic, overlooking the context in which words were used. For example, the word "crisis" may have negative connotations in general but could be used in a positive context, such as "the company successfully averted a crisis."

The advent of ML techniques brought significant advancements in sentiment analysis. ML models, trained on labeled financial news articles, were capable of learning from the context in which words were used and thus provided more accurate sentiment classifications \citep{schumaker2009textual}. Nevertheless, traditional ML techniques, such as Naive Bayes, support vector machines, and decision trees, often failed to capture long-term dependencies in the text and were susceptible to the high dimensionality of the text data \citep{chen2018textual}.

However, deep learning (DL) techniques and their application to NLP has significantly evolved sentiment analysis. This is particularly evident with the emergence of advanced language models (LMs) such as ELMo \citep{peters2018deep}, ULMFit \citep{howard2018universal}, and transformer-based architectures like BERT, which exhibit superior performance in sentiment classification tasks. With their ability to comprehend context, these models can capture long-term dependencies in text and effectively reduce data dimensionality via high-quality embeddings. Domain-specific adaptations of BERT, like FinBERT, further extend these capabilities, specifically tuned for financial texts, hence bolstering the accuracy and reliability of financial sentiment analysis. These developments highlight the transformational potential of LMs in the field, enabling a more nuanced understanding and prediction of sentiments.

Despite recent advancements, there are inherent challenges in financial sentiment analysis. These include not only the need to comprehend domain-specific jargon and to disentangle subtle sentiments associated with multiple financial instruments within the same context, but also the ability to adapt the sentiment output based on specific use-cases, prior conditions or topics. For instance, a sentence might be perceived as negative for the general market economy but simultaneously positive for a specific stock. Current models, such as FinBERT, primarily consider the perspective of the experts that annotated the training data, limiting their flexibility to adjust their output based on these finer nuances and making such framing of sentiment context-dependent an open challenge. This highlights the necessity for future research directions that enhance the adaptability and performance of sentiment analysis in finance. In this regard, the advent of LLMs like GPT offers promising potential. The inherent capabilities of these models, including contextual understanding and adaptability, could provide valuable solutions to these challenges, opening new avenues for exploration and study in this domain \citep{radford2019language}.

\subsection{ChatGPT and Related AI Tools}

With the success of GPT, an array of LLMs have been developed that showcase remarkable performance in various NLP tasks. Each of these models brings unique advantages and capabilities to the field of NLP, including financial applications.

Among these, BloombergGPT stands out as a prominent example that has exhibited excellent performance in several financial NLP tasks \citep{wu2023bloomberggpt}. Designed by Bloomberg's AI team, this model has been trained on a massive corpus of financial texts. However, despite its notable success in various financial tasks, BloombergGPT is yet to provide an open API, with its application being primarily internal to Bloomberg as of May 2023.

Another noteworthy mention in the realm of LLMs is Google's Bard the main rival of ChatGPT. This AI conversational service is powered by Google’s LAMDA (Language Model for Dialogue Applications), combining architectural features of BERT and GPT \citep{thoppilan2022lamda}. Despite its significant potential for engaging and contextually aware dialogues, it shares a similar limitation with BloombergGPT; as of the time of writing, there is a widely open API available for Bard.

An open-source alternative to GPT-3, BLOOM \citep{scao2022bloom}, has also gained recognition among prominent LLMs. BLOOM, though open-source, poses its own challenges. To effectively utilize BLOOM, a considerable amount of specialized knowledge is required, alongside sufficient computational resources. Furthermore, there isn't a tuned version available for conversation-specific tasks, a feature that makes models like ChatGPT particularly useful.

Post the advent of ChatGPT, numerous LLMs with a focus on more specific tasks have emerged, such as those designed for code completion \citep{dakhel2023github}, content creation, and marketing applications. These models, although more narrow in their application, bring additional dimensions of utility and specialization to the table, further expanding the possibilities and impacts of LLMs. However the remarkable advancement and proliferation of large language models, ChatGPT maintains a leading position within the field \citep{jasper23report}. Its accessibility through an open API, vast scale of training data, and applicability across an array of tasks highlights its considerable potential and practical utility.

Notwithstanding the advancements in the application of ChatGPT across various domains such as healthcare and education \citep{sallam2023chatgpt}, its explicit application to financial sentiment analysis remains an underexplored area. Interestingly, there seems to be a paucity of studies directly evaluating the performance of ChatGPT through its API, which is a prerequisite for broader use and integration into third-party applications. No known studies have, as yet, evaluated the performance of ChatGPT specifically in the domain of financial sentiment analysis, and particularly within the context of the forex market. This presents a remarkable gap in the current body of research that this present study seeks to address. By casting our investigative lens on the application and performance of ChatGPT within the sphere of sentiment analysis, and by directly interacting with its API, we aim to contribute novel insights and findings to this emergent field. This underlines the real-world applicability and practicality of our approach, as it aligns with the deployment scenario of incorporating ChatGPT into applications and services.

\section{Methodology}

In this section, we describe our methodology, focusing on evaluating ChatGPT's performance in financial sentiment analysis. We discuss data collection, preprocessing, and the deployment of ChatGPT using its API. Details about the experimental setup, including prompt design and evaluation criteria, are also presented, with an emphasis on the practical implications of our findings in the financial sector.

\label{sec:3}
\subsection{Dataset Creation and Annotation}
\label{sec:3.1}
\begin{figure}[ht]
    \centering
        \includegraphics[width=\linewidth]{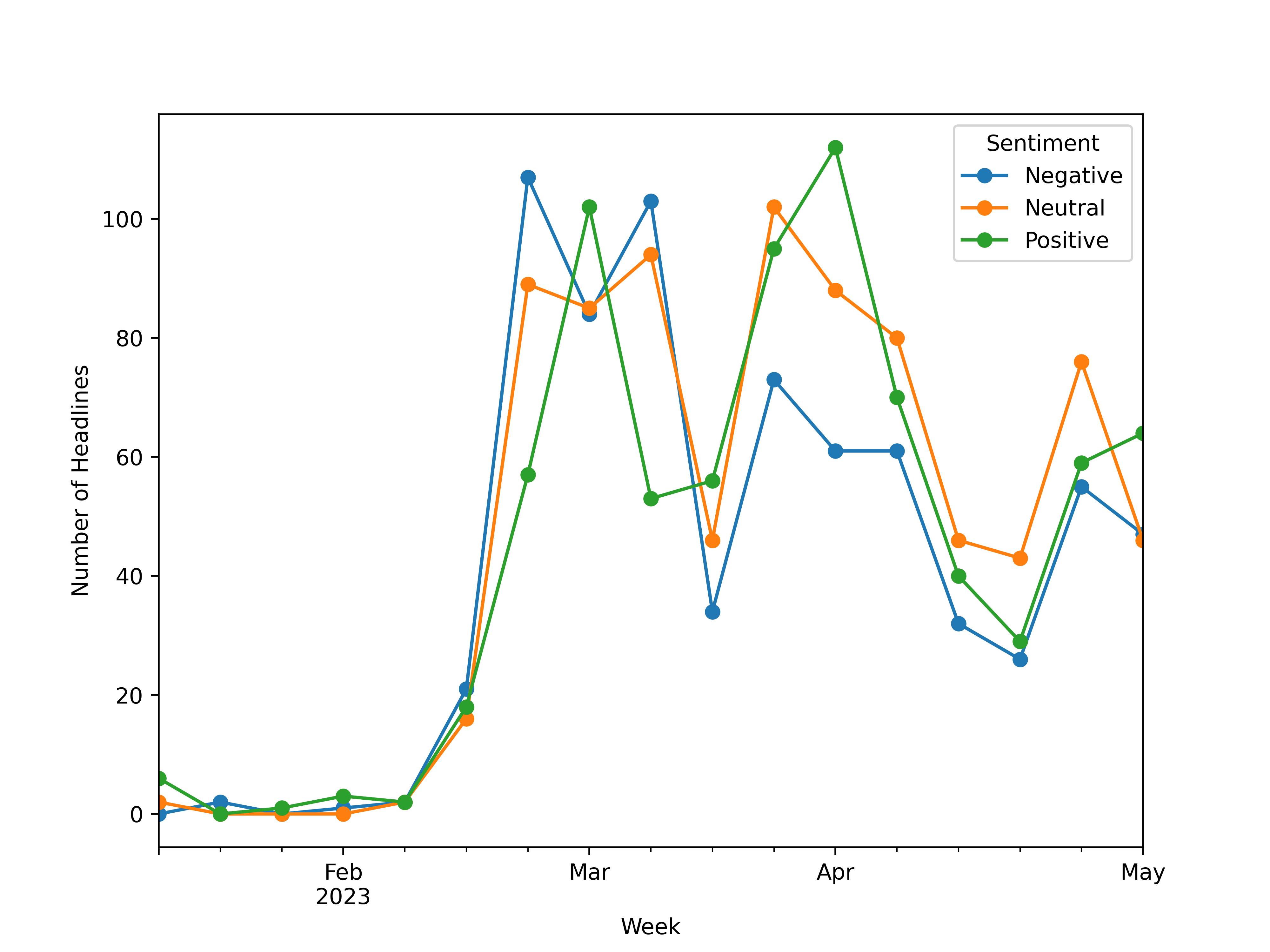}
        \caption{Time Series Plot of Sentiment Labels per Week}
        \label{fig:d3}
\end{figure}

For a comprehensive exploration in this study, we collected news headlines relevant to key forex pairs: AUDUSD, EURCHF, EURUSD, GBPUSD, and USDJPY. The data was extracted from reputable platforms Forex Live\footnote{https://www.forexlive.com/} and FXstreet\footnote{https://www.fxstreet.com/} over a period of 86 days, from January to May 2023 (Fig. \ref{fig:d3}). According to Similarweb \citep{Similarweb}, these platforms are recognized as leading sources for reliable news and current forex analyses, offering critical insights into significant economic and political events.

\begin{figure}[ht]
    \centering
        \includegraphics[width=\linewidth]{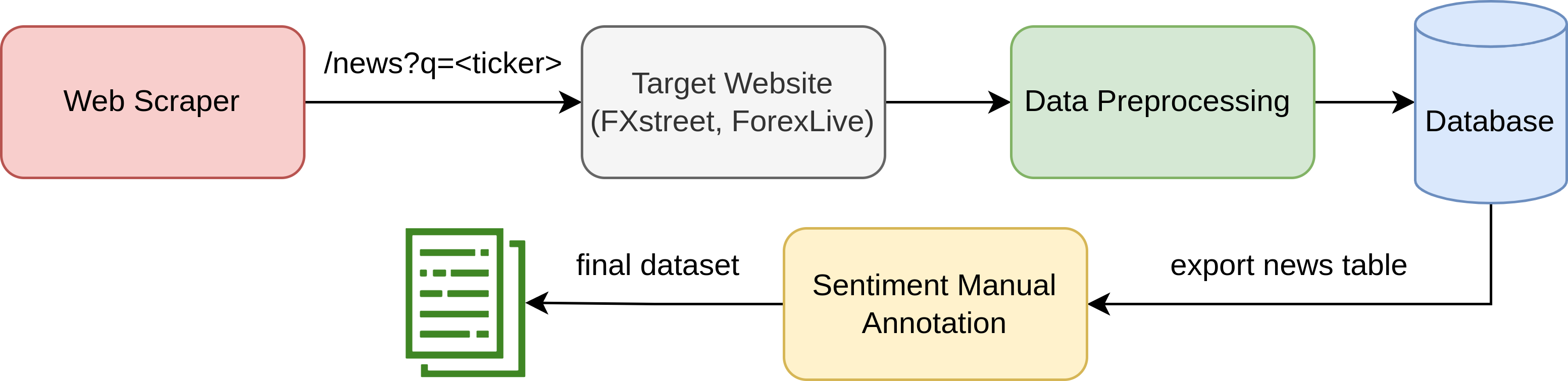}
        \caption{Dataset Creation Process}
        \label{fig:scraper}
\end{figure}

Our dataset comprises 2,291 unique news headlines. Each headline includes an associated forex pair, timestamp, source, author, URL, and the corresponding article text. Data collection was accomplished using web scraping techniques executed via a custom service on a virtual machine. This service periodically retrieves the latest news for a specified forex pair (ticker) from each platform, parsing all available information. The collected data is then processed to extract necessary details such as the article's timestamp, author, and URL. The URL is further used to retrieve the full text of each article. Finally, the data are stored in a database (Fig. \ref{fig:scraper}). This data acquisition process repeats approximately every 15 minutes.

\begin{figure}[ht]
    \centering
        \includegraphics[width=\linewidth]{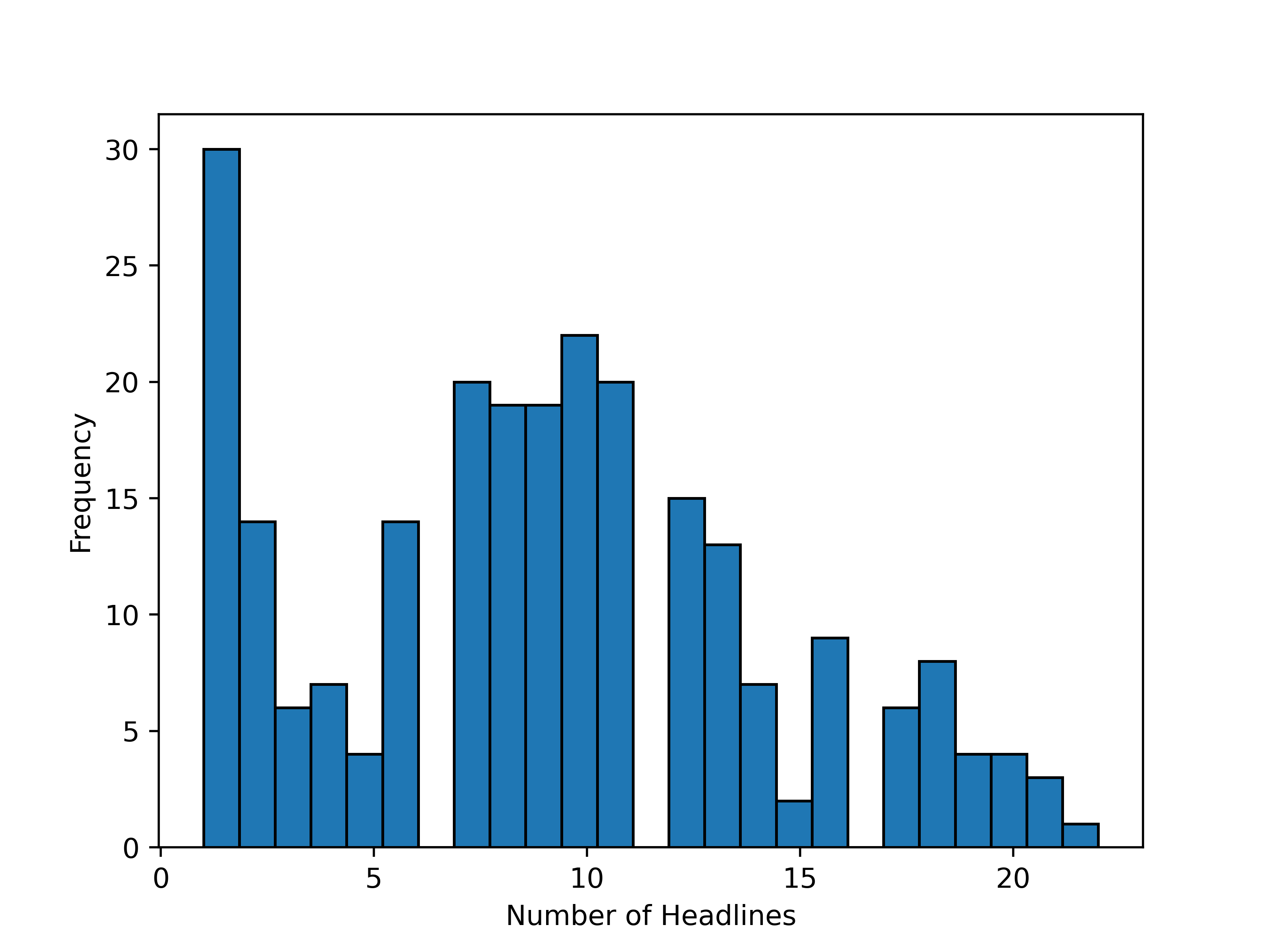}
        \caption{Histogram of News Volume per Day }
        \label{fig:data_vol}
\end{figure}

To ensure the reliability of the dataset and the validity of the experimental evaluations, we manually annotated each headline for sentiment. Instead of solely focusing on the textual content, we ascertained sentiment based on the potential short-term impact of the headline on its corresponding forex pair. This method recognizes the currency market's acute sensitivity to economic news, which significantly influences many trading strategies \citep{evans2005currency}. As such, this dataset could serve as an invaluable resource for fine-tuning sentiment analysis models in the financial realm.

\begin{figure}[ht]
    \centering
        \includegraphics[width=\linewidth]{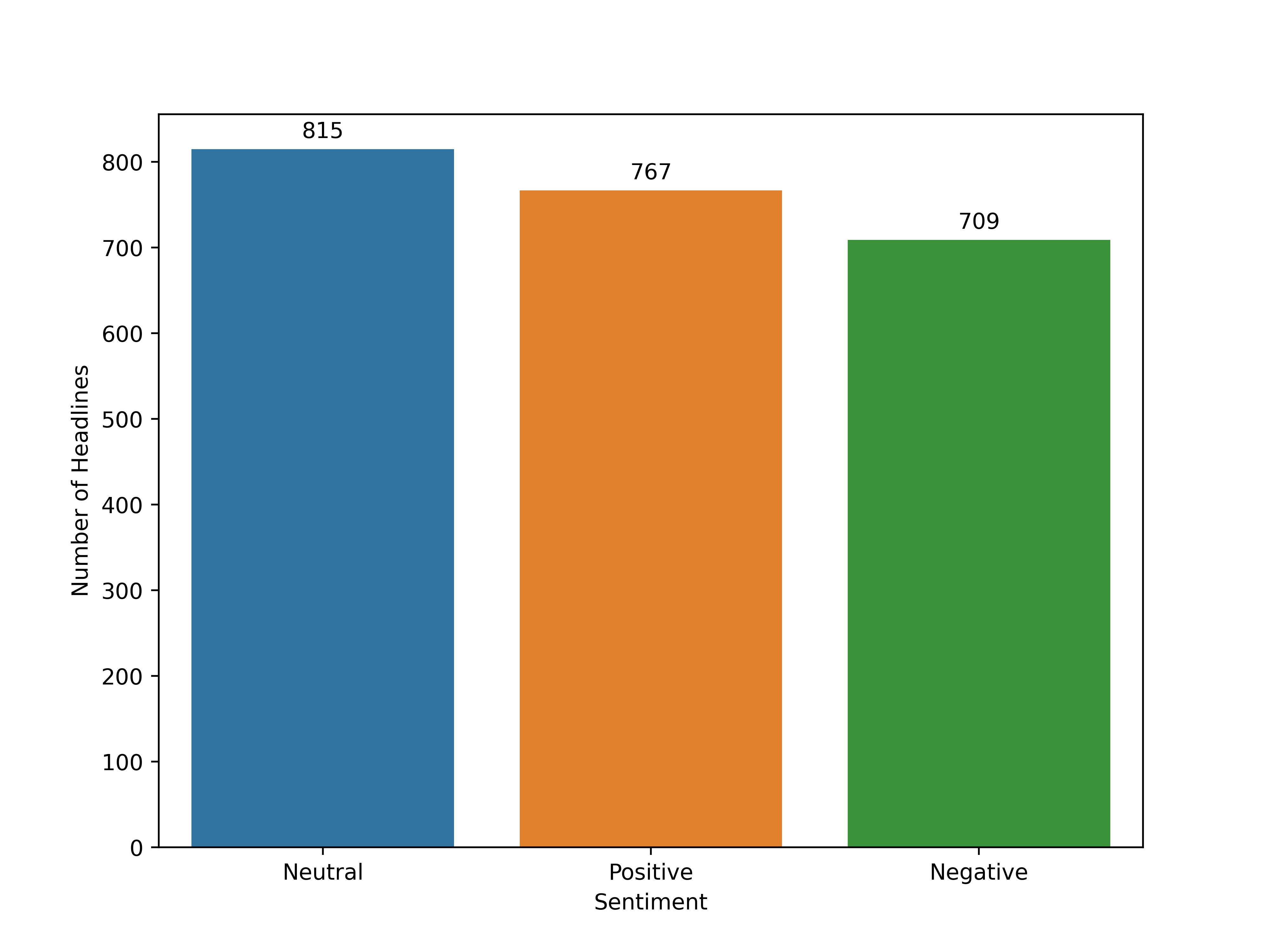}
        \caption{Sentiment Distribution}
        \label{fig:d1}
\end{figure}

\begin{figure}[ht]
    \centering
        \includegraphics[width=\linewidth]{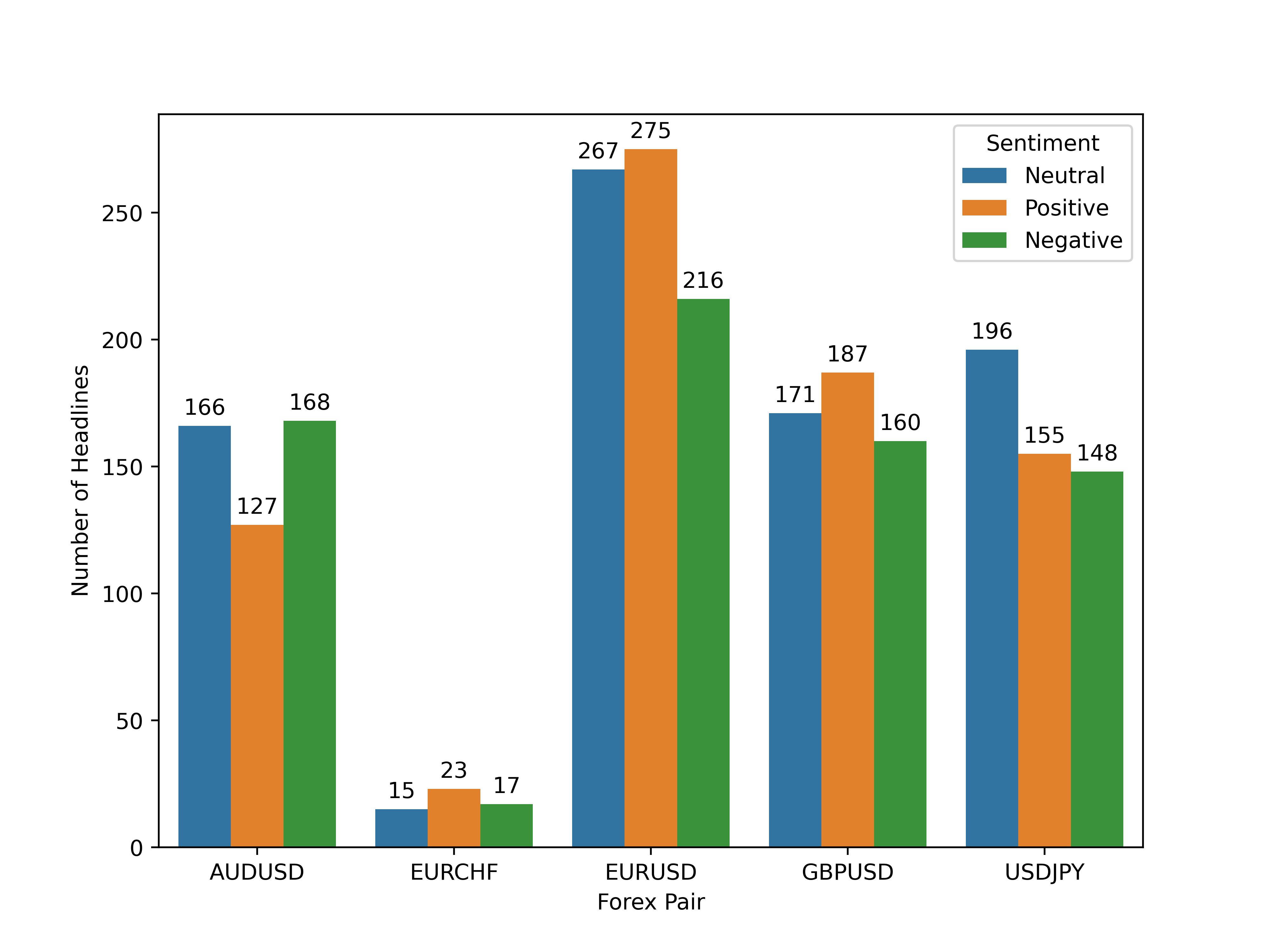}
        \caption{Sentiment Distribution per Forex Pair}
        \label{fig:d2}
\end{figure}

\begin{figure}[ht]
    \centering
        \includegraphics[width=\linewidth]{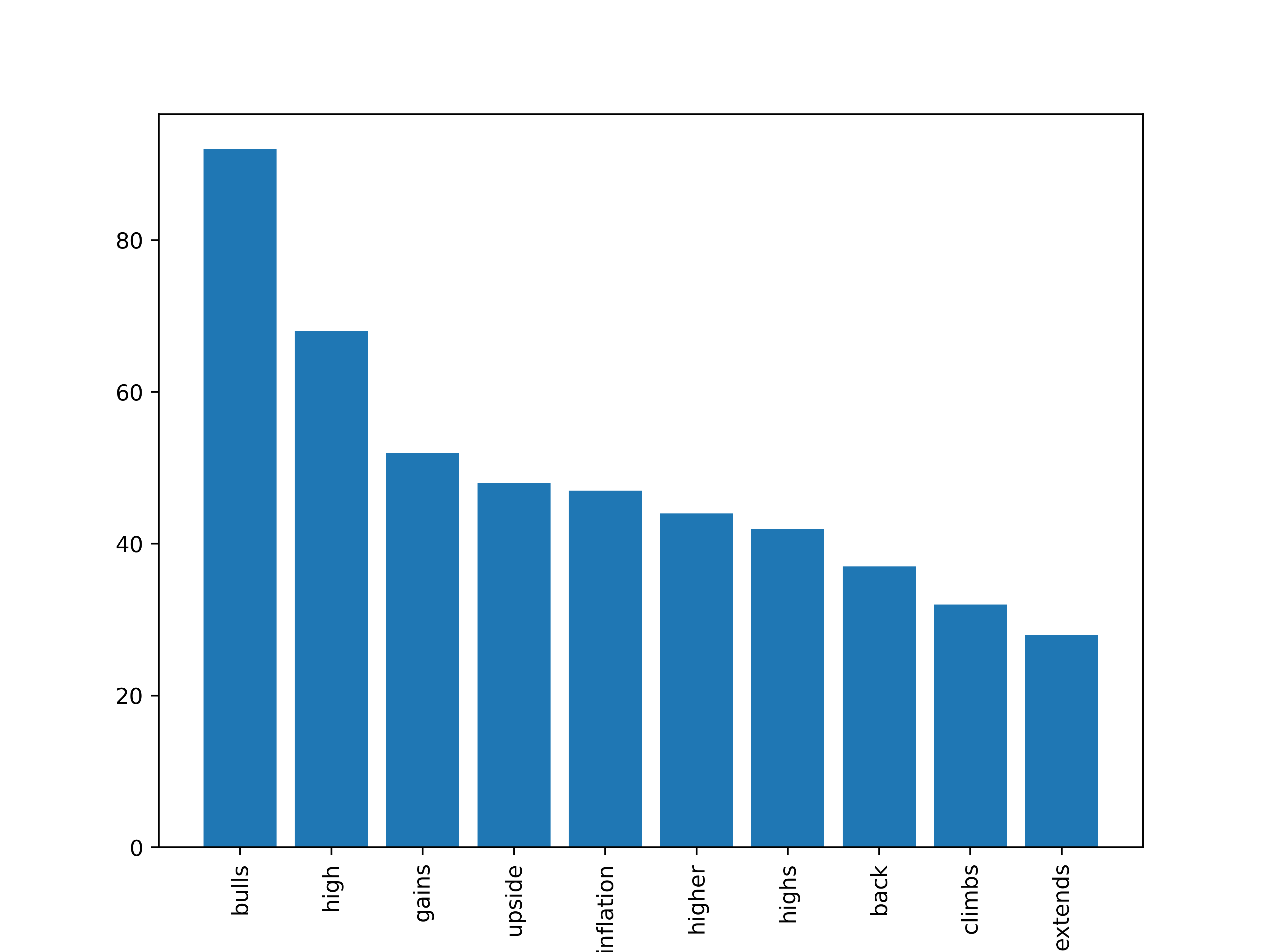}
        \caption{Most Common Tokens on Positive Headlines}
        \label{fig:d4}
\end{figure}

\begin{figure}[ht]
    \centering
        \includegraphics[width=\linewidth]{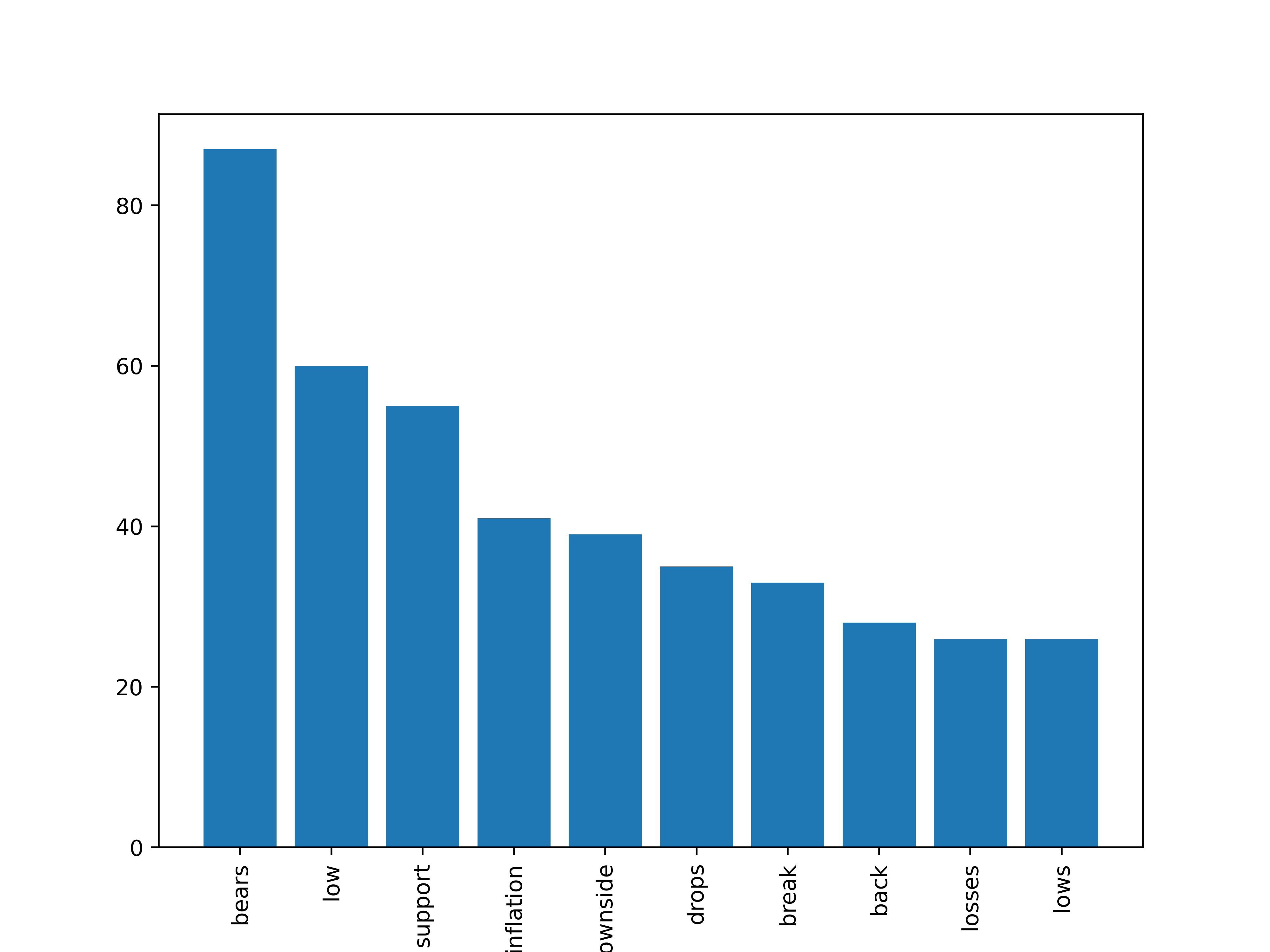}
        \caption{Most Common Tokens on Negative Headlines}
        \label{fig:d5}
\end{figure}

We used three categories for annotation: 'positive', 'negative', and 'neutral', which correspond to bullish, bearish, and hold sentiments respectively, for the forex pair linked to each headline. Table \ref{tab:1} provides examples of annotated headlines along with brief explanations of the assigned sentiment. Fig. \ref{fig:d1} and \ref{fig:d2} represent the overall sentiment distribution within the dataset, as well as the distribution for each individual forex pair. Furthermore, Fig. \ref{fig:d4} and \ref{fig:d5} display the most common tokens in the headlines categorized as positive and negative sentiment, after stopwords, forex pairs, and central banks' names have been removed. Table \ref{tab:tokens} provides further quantitative detail on the distribution of tokens in our dataset. It presents the average and standard deviation of the number of tokens in both the complete article and the headlines. The average number of tokens in each article text is approximately 263, with a standard deviation of 159. For headlines, the average number of tokens is around 12, with a standard deviation of 3. These statistics reflect the variability in the length of texts and headlines in the dataset.

Importantly, the number of tokens plays a significant role in the response latency, cost, and performance of language models like ChatGPT. Longer texts require more computational resources and time for processing, which can lead to increased latency and cost. Conversely, shorter texts can be processed faster and at a lower cost, but they might not provide as much contextual information for the model to generate accurate and relevant responses. 
\begin{table*}[htbp]
\small
\centering
\caption{Examples of Annotated Headlines}
\label{tab:1}
\begin{tabular}{p{1.5cm}p{6cm}p{1.5cm}p{7.5cm}}
\hline
\textbf{Forex Pair} & \textbf{Headline} & \textbf{Sentiment} & \textbf{Explanation} \\
\hline
GBPUSD & Diminishing bets for a move to 12400 & Neutral & Lack of strong sentiment in either direction\\
GBPUSD & No reasons to dislike Cable in the very near term as long as the Dollar momentum remains soft  & Positive & Positive sentiment towards GBPUSD (Cable) in the near term\\
GBPUSD & When are the UK jobs and how could they affect GBPUSD & Neutral & Poses a question and does not express a clear sentiment\\
USDJPY & BoJ’s Ueda: Appropriate to continue monetary easing to achieve 2\% inflation target with wage growth & Positive & Monetary easing from Bank of Japan (BoJ) could lead to a weaker JPY in the short term due to increased money supply\\
USDJPY & Dollar rebounds despite US data. Yen gains amid lower yields & Neutral & Since both the USD and JPY are gaining, the effects on the USDJPY forex pair might offset each other\\
USDJPY & USDJPY to reach 124 by Q4 as the likelihood of a BoJ policy shift should accelerate Yen gains & Negative & USDJPY is expected to reach a lower value, with the USD losing value against the JPY.\\
AUDUSD & RBA Governor Lowe’s Testimony High inflation is damaging and corrosive & Positive & Reserve Bank of Australia (RBA) expresses concerns about inflation. Typically, central banks combat high inflation with higher interest rates, which could strengthen AUD.\\
\hline
\end{tabular}
\end{table*}

\begin{table}[htbp]
\centering
\small
\caption{Mean and Standard Deviation of the Number of Tokens}
\label{tab:tokens}
\begin{tabular}{lcc}
\hline
& \textbf{Mean} & \textbf{Standard Deviation} \\
\hline
\textbf{Headline Tokens} & 11.69 & 2.82 \\
\textbf{Article Tokens} & 262.58 & 158.99 \\
\hline
\end{tabular}
\end{table}

In an effort to contribute to the research community and foster transparency in our methodology, we have made our curated dataset publicly accessible. This collection, featuring news headlines linked to major forex pairs alongside their respective sentiments, is among the first of its kind. We anticipate that it will serve as an invaluable resource for researchers and practitioners keen on exploring or applying machine learning techniques, specifically within the scope of financial sentiment analysis.

We hope that this contribution will spur further research in this field, enriching our shared comprehension of the dynamic interplay between financial news sentiment, AI models, and market behavior. Detailed information about how to access and use the dataset is available at Zenodo repository \citep{fxdataset}.

\subsection{Establishing Baseline: Sentiment Classification using FinBERT}

\label{sec:3.2}

To validate the effectiveness of our proposed approach, we contrast ChatGPT's performance with a conventional benchmark in financial sentiment analysis—FinBERT. Selected for its exceptional proficiency in classifying sentiments expressed in financial texts, FinBERT provides a robust baseline for comparing and evaluating our results. By doing so, we aim to demonstrate the substantial potential of ChatGPT in this specialized task.

In terms of implementation, we utilized the Transformers library from Hugging Face to operate FinBERT \citep{wolf2020transformers}. This Python-based API provides an intuitive interface for employing pre-trained transformer models. By leveraging the convenience and ease of use of the Transformers library, we efficiently integrated FinBERT into our experiments. This high-level process involved loading the FinBERT model and tokenizer from Hugging Face's model repository, tokenizing the headlines, and feeding them into the model. The FinBERT model subsequently outputs a set of probabilities for each sentiment class (positive, negative, and neutral), representing the model's confidence in associating the input headline with each sentiment category. These probabilities are used to determine the predicted class and a sentiment score (notated as FinBERT-N) for each headline. The sentiment score is computed by subtracting the negative class probability from the positive class probability.

\subsection{Exploring Sentiment Classification with ChatGPT}

The primary goal of this work is to explore the potential of ChatGPT in the field of financial sentiment analysis. Given its extensive training on diverse texts, ChatGPT is expected to demonstrate an advanced understanding of context, a key factor in interpreting financial news sentiment. The rationale for utilizing ChatGPT lies in its potential ability to comprehend not only the literal meanings of words but also the underlying implications, idioms, or sentiments that can be industry-specific or topic-centric. For example, in the forex market, a piece of news might have a positive sentiment for one currency (e.g., JPY) but a negative sentiment for the corresponding forex pair (e.g., USDJPY). This nuanced contextual understanding, crucial to most sentiment or emotion analysis tasks, is where ChatGPT could potentially outshine established LMs such as BERT.

To evaluate the capabilities of ChatGPT in financial sentiment analysis, we devised and conducted a series of experiments through a zero-shot prompting approach. 
With zero-shot prompting, we harness ChatGPT's pre-existing training and understanding of language patterns and context to generate desired responses without any task-specific fine-tuning or further training. This approach allows us to directly evaluate ChatGPT's inherent capabilities in sentiment analysis without any additional modifications, demonstrating the potential for direct integration and utilization of ChatGPT in various applications.

Table \ref{tab:2} details the prompts used for sentiment classification. These prompts range from emulating the perspectives of a financial analyst evaluating news (GPT-P1 and P2) to a sentiment analysis AI model (such as FinBERT) assessing a headline (GPT-P3), a forex trader reacting to market updates (GPT-P4), as well as a prompt evaluating all the available daily news per forex pair (GPT-P5) and all available daily news (GPT-P6). Furthermore, we implemented variations of these prompts requiring numerical sentiment output in the form of a sentiment score (GPT P1N-P6N) ranging from $[-1,1]$ instead of a sentiment class label. Prompts GPT-P6 and P6N were designed to ask ChatGPT to process a list of headlines and return a collective sentiment in a structured format, specifically JSON. This feature evaluates ChatGPT's capability to integrate seamlessly into existing services by producing easily consumable output. Such a capability is vital in real-world applications where the model's results need to be immediately utilized by other services or systems.

\begin{table*}[htbp]
\small
\centering
\caption{Experimental ChatGPT Prompts for Sentiment Classification}
\label{tab:2}
\begin{tabular}{p{2cm}p{15cm}}
\hline
\textbf{Prompt Abbr.} & \textbf{Prompt} \\
\hline
GPT-P1 & Act as a financial expert holding \{ticker\}. How do you feel about the headline \{headline\}? Answer in one token: positive, negative, or neutral. \\
GPT-P2 & Act as a financial expert. Classify the sentiment for \{ticker\} based only on the headline \{headline\}. Answer in one token: positive, negative, or neutral.  \\
GPT-P3 & Act as a sentiment analysis model trained on financial news headlines. Classify the sentiment of the headline \{headline\}. Answer in one token: positive, negative, or neutral.  \\
GPT-P4 & Act as an expert at forex trading holding \{ticker\}. Based only on the headline \{headline\},  will you buy, sell or hold \{ticker\} in the short term? Answer in one token: positive for buy, negative for sell, or neutral for hold position.  \\
GPT-P5 & Act as an expert at forex trading holding \{ticker\}. Based only on the following list of headlines \{ticker\_daily\_headlines\}, will you buy, sell or hold \{ticker\} in the short term? Answer in one token: positive for buy, negative for sell, or neutral for hold position  \\
GPT-P6 & Act as a sentiment analysis service of a financial platform. Based only on the following list of headlines \{all\_daily\_headlines\}, provide a summary of the daily sentiment for the forex pairs: \{tickers\}. Provide only the sentiment per forex pair in JSON format eg \{'USDJPY': 'positive', 'EURUSD': 'neutral'\}. The sentiment can be positive for buy, negative for sell, or neutral for hold position.\\
\hline
\end{tabular}
\end{table*}

To conduct these sentiment analysis experiments, we leveraged the GPT-3.5-turbo model available from OpenAI's Python library \citep{brown2020language}. Each prompt was fed to the model as a single user instruction. Depending on the prompt, we fine-tuned several parameters to optimize the model's output. Specifically, the \textit{max\_tokens} parameter in GPT P1-P5 was set to 1 to constrain the model's response to a single token. For prompts GPT P1N-P4N, P5N, P6-P6N, the \textit{max\_tokens} parameter was set to 10, 20, and 200 tokens respectively, according to their expected response length. Consequently, dedicated processing was performed on the response text to extract the necessary information per prompt. For instance, we crafted a straightforward function with a regular expression to extract the JSON (JavaScript Object Notation) from P6's response. The \textit{temperature} parameter was set to 0.2 to guide the model towards generating more deterministic outputs. It's worth noting that we ended up with these prompts after extensive prompt engineering to ensure that GPT includes the desired information (i.e., sentiment class or score) in its responses. 

For each prompt, we monitored the time taken to receive a response from the API to track the task latency and evaluate GPT's efficiency. We also logged the number of tokens used in the completion and the number of tokens in the prompt for cost management and resource usage tracking.

Each experiment was run iteratively over the forex dataset presented in Section \ref{sec:3.1}. To ensure the smooth execution of the iterative process, we incorporated error handling mechanisms to address potential API exceptions or response errors. It's essential to note that running these sentiment analysis experiments over large datasets consumes API tokens and incurs costs in line with OpenAI's pricing structure. However, the insights obtained from the sentiment analysis greatly outweigh the computational expenses, offering invaluable intelligence for enhanced decision making in finance.

\subsection{Evaluation Metrics}

The evaluation approach for sentiment classification in this study was twofold, encompassing both a traditional evaluation grounded in the comparison with the true sentiment class and a market-related model evaluation. The former method focused on the models that predicted a sentiment class for each headline (i.e., FinBERT and GPT's P1-P4), by comparing the predicted sentiment class with the actual sentiment class as annotated in our dataset. On the other hand, the market-related model evaluation was applicable to all considered models, achieved by assigning the corresponding integer code to the predicted sentiment classes. Specifically for FinBERT, the sentiment score was derived from the predicted probabilities per class, as detailed in Section \ref{sec:3.2}.

\subsubsection{Evaluation Based on True Class}

To gauge the effectiveness of our sentiment classification models, we devised an evaluation framework rooted in multiple performance metrics, such as accuracy, precision, recall, and F1-score. These metrics illuminate the extent to which the models accurately predict sentiment classes defined in our annotated dataset.

A key metric in our evaluation toolbox is the Sentiment Mean Absolute Error (S-MAE) expressed by Eq. \ref{eq:1}, where $N$ represents the total number of instances or data points, $y_i$ denotes the true sentiment class for the i-th instance, and $\hat{y}_i$ represents the predicted sentiment class for the i-th instance. This metric quantifies the mean absolute difference between the integer-represented true sentiment classes and the predicted sentiment classes, with -1, 0, and 1 representing negative, neutral, and positive sentiments, respectively.

\begin{equation}
S-MAE = \frac{1}{N} \sum_{i=1}^{N} |y_{i} - \hat{y}_{i}|
\label{eq:1}
\end{equation}

The rationale behind utilizing S-MAE is its capacity to distinctly penalize models that significantly misjudge sentiment classification. This aspect is particularly important given the intricate nature of sentiment analysis in financial news, where sentiment interpretation can be complex and nuanced. For instance, a headline with a positive sentiment towards a particular currency (e.g., USD) could suggest a negative sentiment for its paired currency (e.g., EURUSD). In another case, a negative sentiment towards the economy due to an announcement from a central bank might convey a positive market sentiment for the corresponding currency. These subtleties underscore the necessity of penalizing not only incorrect polarity predictions but also off-target intensity estimations.

Moreover, the evaluation includes a comparative analysis among different models. Specifically, we compare the performance of the FinBERT model with that of ChatGPT under varying prompts (P1-P4). This approach intends to highlight the impact of prompt selection on the performance of ChatGPT, offering a nuanced understanding of how these different models perform in sentiment classification tasks.

\subsubsection{Model Evaluation in Relation to the Market}

Aside from conventional classification metrics, we also evaluated the models' output in connection with actual market returns. This evaluation strategy is driven by the primary intended application of these models - to guide trading decisions in financial markets.

For this component of the evaluation, we aggregate the sentiment scores for each forex pair on a daily basis. In particular, the sentiment scores of all headlines concerning a specific forex pair within a single day are summed to derive a daily sentiment score. This score is then compared with the actual daily return of the corresponding forex pair using Pearson correlation. This correlation is computed to assess whether the sentiment scores, as predicted by the models, have any relation to real market movements.

In addition to correlation with market returns, the models are also evaluated on their ability to accurately predict the direction of market movement, a critical element in financial decision-making \citep{blaskowitz2011economic}. This metric, termed Directional Accuracy (DA), is computed as the percentage of days where the predicted sentiment direction (positive/negative) aligns with the actual market return direction. For example, if the model correctly predicted the direction of market movement on 65 days out of 100, then the Directional Accuracy of the model would be 65\%.

For the purpose of this market-related evaluation, the sentiment score for prompts GPT-P1 to P6 is regarded as an integer value of the predicted class. This interpretation allows the sentiment score to be quantified and aggregated in a way that aligns with its intended meaning - a measure of the overall daily sentiment towards a specific forex pair.

This evaluation includes all models (FinBERT, GPT P1-P6, and P1N-P6N) and, by combining Directional Accuracy with Pearson correlation, offers a more holistic view of the models' performance in a financial context.

\section{Results}
\label{sec:4}
This section details the findings obtained from our extensive analysis and experiments, aimed at scrutinizing the effectiveness of ChatGPT in financial sentiment analysis. We present a comprehensive discussion of these results, comparing the performance of the various ChatGPT prompts and the established baseline, FinBERT, across various metrics. 

\subsection{Sentiment Classification Performance}

Table \ref{tab:r1} showcases the sentiment classification performance of various models including FinBERT and distinct prompts from the ChatGPT (GPT-P1 to P4) used in our dataset. The evaluation metrics include Accuracy, Precision, Recall, F1-Score, and S-MAE.

As presented in the table, GPT models consistently outperform FinBERT across all metrics. FinBERT, which is a general-purpose language model trained on financial text, shows an accuracy of 0.561, a precision of 0.560, and a recall of 0.562. Its F1-Score stands at 0.556 and it yields an S-MAE of 0.540. Although these figures are decent, they are significantly overshadowed by the performance of the GPT models. Interestingly, GPT-P3, which mirrors the FinBERT model in acting as a sentiment analysis model trained on financial news headlines performs relatively well. However, it doesn't match the performance of other GPT models that leverage the forex pair information in their prompts.

GPT-P1, which is designed to function as a financial expert performing emotion analysis on headlines while taking into account the related forex pair, also registers a commendable performance in the sentiment classification task. However, it underperforms slightly compared to GPT-P2, with its metrics trailing by approximately 8\%, indicating that while the emotion-based approach of P1 is effective, the sentiment-oriented approach of P2 appears to offer more precision in this context.

GPT-P2, and P4, which act as financial experts considering the related forex pair in their sentiment analysis, register substantial improvements in the sentiment classification task. GPT-P2 leads the pack with the highest accuracy, recall, and F1-Score around 0.790. GPT-P4, designed to impersonate a forex trader making trading decisions based on the headline sentiment, posts similar performance metrics to GPT-P2, despite the different framing of the prompt. This aligns with how the dataset was annotated, indicating that role-playing prompts, such as those used in GPT-P4 and GPT-P2, are especially effective.

\begin{table}[htbp]
\small
\centering
\caption{Performance Results in Sentiment Classification}
\label{tab:r1}
\begin{tabular}{lccccc}
\hline
\textbf{Model} & \textbf{Accuracy} & \textbf{Precision} & \textbf{Recall} & \textbf{F1} & \textbf{S-MAE} \\
\hline
FinBERT & 0.561 & 0.560 & 0.562 & 0.556 & 0.540 \\
GPT-P1 & 0.730 & 0.760 & 0.730 & 0.725 & 0.300 \\
GPT-P2 & \textbf{0.790} & 0.797 & \textbf{0.790} & \textbf{0.790} & 0.227 \\
GPT-P3 & 0.735 & 0.780 & 0.735 & 0.737 & 0.282 \\
GPT-P4 & 0.784 & \textbf{0.804} & 0.784 & 0.789 & \textbf{0.221} \\
\hline
\end{tabular}
\end{table}

Furthermore, as depicted in Table \ref{tab:r2}, the performance of the GPT models varies depending on the specific forex pair. For the AUDUSD pair, GPT-P2 demonstrates superior performance in terms of accuracy, recall, F1-score, and S-MAE, while GPT-P4 outperforms the other models in precision. In the case of the USDJPY pair, GPT-P2 dominates all performance metrics. The evaluation of the EURCHF pair reveals a split performance dominance, with GPT-P2 leading in accuracy, recall, and F1-score, while GPT-P4 stands out in precision and S-MAE. Notably, GPT-P4 dominates in all performance metrics for both the EURUSD and GBPUSD pairs, underlining the effectiveness of GPT-P4 for sentiment analysis specifically associated with these forex pairs. 

\begin{table}[htbp]
\small
\centering
\caption{Best Performing Model in Sentiment Classification per forex pair}
\label{tab:r2}
\begin{tabular}{lccccc}
\hline
\textbf{FX Pair} &  \textbf{Accuracy} & \textbf{Precision} & \textbf{Recall} &   \textbf{F1} &  \textbf{S-MAE} \\
\hline
AUDUSD &  P2 &  P4 &  P2 &  P2 &  P2 \\
EURCHF &  P2 &  P4 &  P2 &  P2 &  P4 \\
EURUSD &  P4 &  P4 &  P4 &  P4 &  P4 \\
GBPUSD &  P4 &  P4 &  P4 &  P4 &  P4 \\
USDJPY &  P2 &  P2 &  P2 &  P2 &  P2 \\
\hline
\end{tabular}
\end{table}

However, it's worth noting that 75.6\% of the headlines in the dataset mention the related forex pair at their beginning. This observation suggests that the performance difference between prompts such as GPT-P1, P2, and P4, which take the forex pair into account, could be even more pronounced when dealing with more complex headlines. To explore this, we further examined the performance of the models specifically on a subset of the data with headlines that do not explicitly mention the associated forex pair. This assessment provides insight into how well each model handles sentiment analysis in a more challenging context, where the pertinent topic (i.e., forex pair) is not directly stated in the headline text. The results from this analysis, presented in Table \ref{tab:r3}, showed that GPT-P4 continued to outperform, achieving the highest position across all metrics. This finding underscores GPT-P4's ability to interpret the implicit context in the headlines.

On the other hand, FinBERT's performance slightly declined when the forex pair was not mentioned directly, with accuracy, precision, recall, F1-score, and S-MAE at 0.543, 0.543, 0.543, 0.538, and 0.539 respectively. GPT-P1, P2, and P3 also experienced a dip in performance compared to the results from the complete dataset. However, even with the drop in performance, GPT-P1, P2, and P3 outperformed FinBERT, underlining the superiority of the GPT models in handling complex scenarios. In the case of GPT-P2, despite a decrease in performance, it still secured high scores with accuracy, precision, recall, F1-score, and S-MAE at 0.711, 0.728, 0.711, 0.714, and 0.326 respectively, emphasizing its overall robustness in sentiment classification.

This analysis highlights the strength of GPT models, particularly GPT-P4, in sentiment analysis tasks. These models' ability to maintain a high level of performance even in more challenging scenarios makes them promising tools for real-world applications. Crucially, these results also emphasize the importance of the prompt in instructing ChatGPT. By thoughtfully crafting prompts based on the specifics of the task at hand, we can guide the model to leverage its knowledge more effectively, thus enhancing its performance in sentiment analysis.

\begin{table}[htbp]
\small
\centering
\caption{Performance Results in Sentiment Classification (filtered data)}
\label{tab:r3}
\begin{tabular}{lccccc}
\hline
\textbf{Model} & \textbf{Accuracy} & \textbf{Precision} & \textbf{Recall} & \textbf{F1} & \textbf{S-MAE} \\
\hline
FinBERT & 0.543 & 0.543 & 0.543 & 0.538 & 0.539 \\
GPT-P1 & 0.663 & 0.715 & 0.663 & 0.672 & 0.410 \\
GPT-P2 & 0.711 & 0.728 & 0.711 & 0.714 & 0.326 \\
GPT-P3 & 0.665 & 0.657 & 0.665 & 0.657 & 0.384 \\
GPT-P4 & \textbf{0.765} & \textbf{0.772} & \textbf{0.765} & \textbf{0.763} & \textbf{0.249} \\
\hline
\end{tabular}
\end{table}

\subsection{Sentiment Score relation to the Financial Market}

The second evaluation approach examines how well the sentiment scores predicted by each model align with market price movements. We compare the sentiment scores of each model against the actual daily returns of the respective forex pairs to understand the potential utility of the models in a trading context. This comparison was done in terms of Pearson correlation and DA as presented in Section \ref{sec:3}.

\subsubsection{Correlation with Market Returns}
The correlation coefficient between sentiment scores and market returns is a key factor in evaluating the effectiveness of the presented models. This correlation is measured using the Pearson correlation coefficient, which ranges from -1 to +1. The higher the correlation value, the more directly related the sentiment scores and market returns are. In this context, a higher positive correlation implies that as sentiment improves, market returns also increase. Consequently, the model with the highest correlation between its sentiment scores and market returns can be considered the most accurate in reflecting the market's response to sentiment changes.

\begin{figure}[ht]
    \centering
        \includegraphics[width=\linewidth]{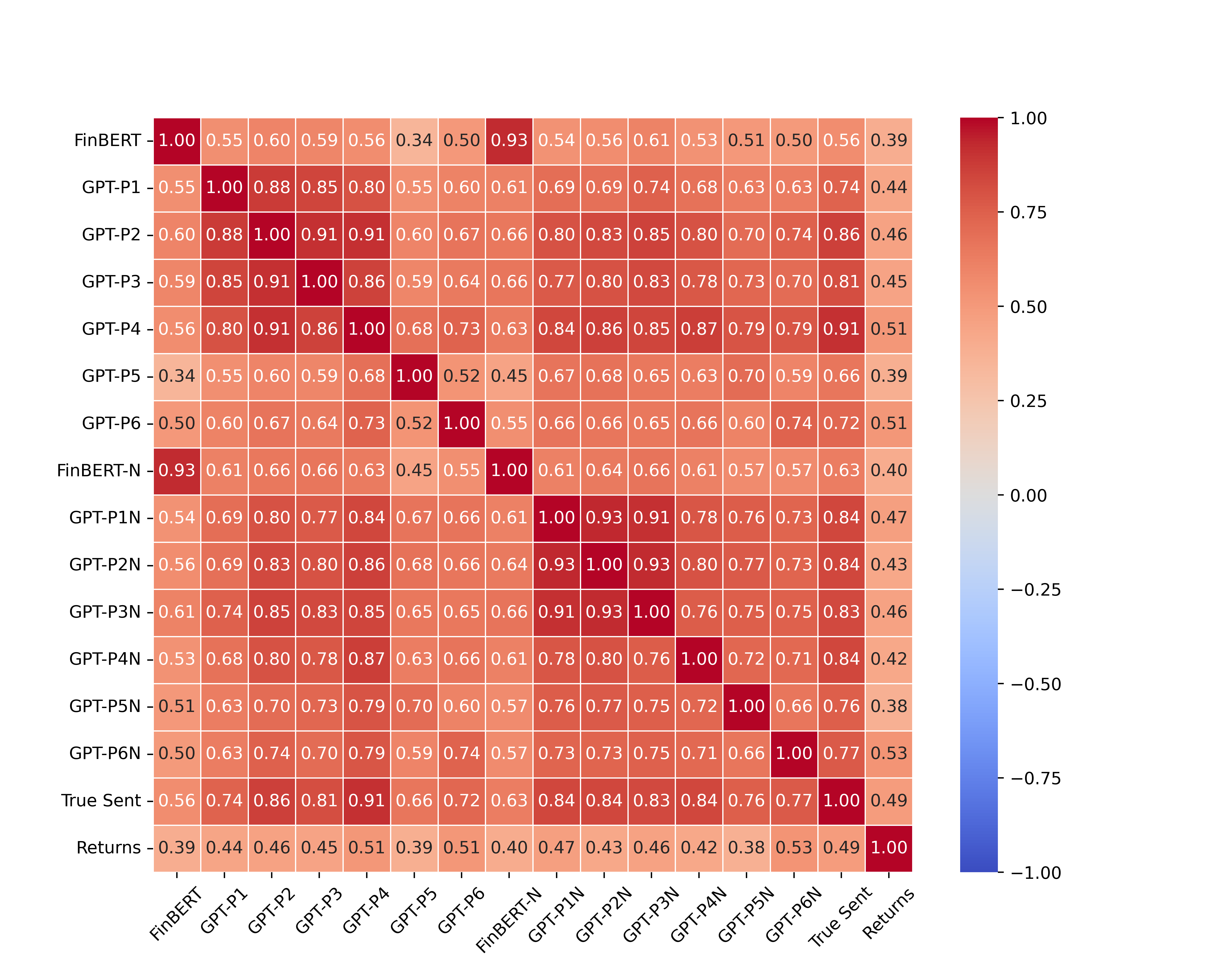}
        \caption{Correlation Matrix of Predicted Sentiment and Market returns}
        \label{fig:res1}
\end{figure}

The correlation matrix represented in Figure \ref{fig:res1} unveils the interplay among various models, the annotated sentiment, and the daily market returns. It is indeed reasonable that GPT-P4 exhibits the highest correlation with the true sentiment, given that P4 mirrors the approach used in annotating the sentiment within the dataset. Similarly, FinBERT correlates the most with GPT-P3 which mimics a sentiment analysis model trained on financial data. However, a more intriguing observation is that GPT-P6N demonstrates a higher correlation with the market returns than the true sentiment itself. This underscores the power of GPT's ability to process all daily news simultaneously, which enhances its capacity to gauge market sentiment with a higher degree of accuracy. Most models generating sentiment scores ranging from -1 to 1 instead of performing sentiment classification exhibit a more fitting alignment with the market movements. This is plausible as the latter models are better positioned to accurately capture news sentiment that is slightly positive or negative. 

It's also worth noting that GPT-P5 showed the lowest correlation with both daily returns and true sentiment, indicating that this prompt may not be as effective in sentiment prediction for market price prediction as the others.

Furthermore, the high correlation between GPT-P1 and P1N, through to GPT-P6 and P6N, suggests that the different versions of the models (categorical and numerical (N) versions) largely agree on the sentiment of the text. This indicates the consistency of the GPT's outputs. 

The correlation between true sentiment and daily returns stands at around 0.49, implying a moderate linear relationship. This suggests that sentiment can explain some but not all of the variations in market prices. Other factors and market dynamics likely come into play as well.

\subsubsection{Directional Accuracy of Sentiment Scores}
The tables (\ref{tab:DA_non_numerical},\ref{tab:DA_numerical}) below presents the DA of different sentiment scoring models. DA reflects the percentage of instances where the predicted and actual direction of change align, with regards to sentiment scores and market movements.

Examining the figures, GPT-P1N is found to have the highest DA at 67.2\%, indicating the superior ability of this model to predict the correct direction of market movement based on sentiment scores. Close contenders are GPT-P3N and GPT-P6N, with DA values around 66\%. Among categorical models, GPT-P5 shows the highest DA, closely followed by GPT-P6 and GPT-P4. However, the numerical models, in general, tend to have a higher DA than their categorical counterparts, underlining their enhanced capacity in capturing the directional changes in market sentiment.

\begin{table}[htbp]
\small
    \centering
    \begin{minipage}{.45\linewidth}
        \centering
        \caption{Directional Accuracy of Non-Numerical Models}
        \label{tab:DA_non_numerical}
        \begin{tabular}{lr}
        \hline
            \textbf{Model} & \textbf{DA} \\
            \hline
            FinBERT & 0.595 \\
            GPT-P1 & 0.583 \\
            GPT-P2 & 0.607 \\
            GPT-P3 & 0.607 \\
            GPT-P4 & 0.640 \\
            GPT-P5 & 0.652 \\
            GPT-P6 & 0.640 \\
            \hline
        \end{tabular}
    \end{minipage}%
    \hspace{0.05\linewidth}%
    \begin{minipage}{.45\linewidth}
        \centering
        \caption{Directional Accuracy of Numerical Models}
        \label{tab:DA_numerical}
        \begin{tabular}{lr}
        \hline
            \textbf{Model} & \textbf{DA} \\
            \hline
            FinBERT-N & 0.599 \\
            GPT-P1N & \textbf{0.672} \\
            GPT-P2N & 0.644 \\
            GPT-P3N & 0.660 \\
            GPT-P4N & 0.611 \\
            GPT-P5N & 0.623 \\
            GPT-P6N & 0.652 \\
            \hline
        \end{tabular}
    \end{minipage} 
\end{table}

Table \ref{tab:DAperModel} showcases the DA of numerical sentiment analysis models, broken down per ticker. Observing the results, GPT-P1N registers the highest DA for AUDUSD and shares the highest DA with GPT-P2N for EURCHF. This indicates the superior ability of GPT-P1N to predict the correct direction of market movement for these particular currency pairs. Meanwhile, for EURUSD and GBPUSD, GPT-P6N outperforms the other models, illustrating its effectiveness in predicting market direction for these tickers. Contrastingly, the FinBERT-N model exhibits the highest DA for USDJPY, while GPT-P6N, which overall is the top performing model, fails capture the sentiment of this specific pair appropriately. This divergence implies that different models might possess distinct strengths in forecasting different market dynamics, suggesting the necessity of a tailored approach for each ticker.

Notably, for the EURCHF currency pair, the DA is generally lower. This might be due to considerable less data being available for these pair, reinforcing the critical role of ample data availability in accurate market direction prediction. Furthermore, it's interesting to observe that GPT-P1N and P3N show the highest DA for this pair, despite exhibiting poorer results compared to GPT-P4N and P6N in all the previous evaluations. Thus, this improved performance for GPT-P1N and P3N might be coincidental. Nevertheless, the results underline the unpredictable nature of financial markets and emphasize the necessity of comprehensive testing on diverse datasets for model validation.

\begin{table}[htbp]
\small
\centering
\caption{Directional Accuracy Results for Each Model per Ticker}
\label{tab:DAperModel}
\begin{tabular}{>{\raggedright\arraybackslash}p{1.3cm}*{5}{>{\centering\arraybackslash}p{1cm}}}
\hline
\textbf{Model} & \textbf{AUDUSD} & \textbf{EURCHF} & \textbf{EURUSD} & \textbf{GBPUSD} & \textbf{USDJPY} \\
\hline
FinBERTN & 0.518 & 0.250 & 0.654 & 0.698 & \textbf{0.714} \\
GPT-P1N & \textbf{0.648} & \textbf{0.500} & 0.673 & 0.679 & 0.696 \\
GPT-P2N & \textbf{0.648} & 0.437 & 0.576 & 0.679 & 0.678 \\
GPT-P3N & 0.630 & \textbf{0.500} & 0.635 & 0.755 & 0.678 \\
GPT-P4N & 0.537 & 0.188 & 0.500 & 0.660 & 0.536 \\
GPT-P5N & 0.593 & 0.469 & 0.615 & 0.660 & 0.589 \\
GPT-P6N & 0.593 & 0.344 & \textbf{0.731} & \textbf{0.792} & 0.607 \\
\hline
\end{tabular}
\end{table}

\subsubsection{Performance and Cost Analysis}

In the context of deploying ML models, particularly in a production environment, performance and cost considerations are crucial. These factors can significantly affect the scalability, efficiency, and overall feasibility of a model. Given this, we evaluate the GPT prompts in terms of processing time and token consumption, both of which directly impact operational costs and performance. Table \ref{tab:model_perf} provides a summary of the average processing time and the number of tokens processed by each prompt. 

\begin{table}[htbp]
\centering
\small
\caption{Average Time and Tokens Processed per ChatGPT Prompt}
\label{tab:model_perf}
\begin{tabular}{lrrrr}
\hline
\textbf{Prompt} & \textbf{Pr. time} & \textbf{Pr. tokens} & \textbf{Hdln. time} & \textbf{Hdln. tokens} \\
\hline
P1 & 0.94 & 63.0 & 0.94 & 63.17 \\
P2 & 0.81 & 67.0 & 0.81 & 67.15 \\
P3 & 0.97 & 66.0 & 0.97 & 65.91 \\
P4 & 0.82 & 84.0 & 0.82 & 84.39 \\
P5 & 0.83 & 193.0 & 0.11 & 24.74 \\
P6 & 5.99 & 592.0 & 0.22 & 22.22 \\
P1N & 2.29 & 90.0 & 2.29 & 90.34 \\
P2N & 1.42 & 85.0 & 1.42 & 84.79 \\
P3N & 1.23 & 80.0 & 1.23 & 79.98 \\
P4N & 1.06 & 91.0 & 1.06 & 90.81 \\
P5N & 1.85 & 201.0 & 0.24 & 25.68 \\
P6N & 6.10 & 600.0 & 0.23 & 22.54 \\
\hline
\end{tabular}
\end{table}

Among the sentiment classification prompts, P1, P2, and P3 exhibit similar average times and tokens. They produce concise responses in less than a second with around 65 tokens. P4, which is the best performing prompt, generates more tokens, which can potentially lead to higher costs. On the other hand, P5 and P6, while generating significantly more tokens, benefit from processing multiple headlines at once, as they have lower average times and tokens per headline, which in addition to higher accuracy, can lead to a cost reduction when processing large quantities of data. The prompts generating numerical outputs follow a similar pattern, but they generally require more time and tokens due to the complexity of generating more accurate responses. 

It is important to note that the processing times mentioned in the study are dependent on many factors and may not be entirely consistent. The load on the ChatGPT model, which can vary depending on the time of day,  other global usage patterns, and end-user location, may significantly impact the processing time. Thus, the average time taken by the models, as presented in the results, while informative for relative comparisons, might not always reflect their real-time performance. Consequently, the time-related results should be interpreted with caution as they might not be entirely reliable.

Taking into account that the cost of utilizing the OpenAI API for the ChatGPT-3.5 model is 0.002 USD per 1K tokens, the financial implications of integrating ChatGPT into existing services seem relatively low, especially considering the volume of data processed daily. As a reference, Bloomberg News produces approximately 5,000 articles on a daily basis \citep{Bloomberg2023}.

To illustrate, let's consider using the P6N model, which generates the highest number of tokens among the models tested in this study. With this model, the daily cost can be calculated as follows: 600 tokens (average generated by P6N per article) * 5,000 articles * 0.002 USD per 1K tokens. This calculation results in a total daily cost of approximately 6 USD.

Hence, even when dealing with substantial data volumes, such as those reported by Bloomberg, the cost of using advanced language models like ChatGPT for sentiment analysis remains reasonably affordable. This affordability, in combination with their demonstrated performance, makes these models a compelling option for real-world financial applications.

However, for applications that require real-time insights and increased accuracy, a mixed approach combining single and multiple news analysis might be necessary. This would allow the flexibility of generating quick and concise responses for individual headlines with P1-P4 (or P1N-P4N), while also processing a larger batch of headlines with P5, P6 (or P5N, P6N) when time allows. 

\section{Discussion}
\label{sec:5}

\subsection{Performance of ChatGPT in Financial Sentiment Analysis}

The results of our study reveal that the ChatGPT model consistently outperforms the FinBERT model in financial sentiment analysis, regardless of the specific prompt used. In particular, prompts GPT-P4, P6, P6N showed considerable performance in sentiment analysis, providing potentially valuable insights for predicting market behavior. This high performance, combined with the model's flexibility through prompt engineering, implies significant potential for fine-tuning and optimizing the application. For example, through strategic framing of the task at hand, as demonstrated by prompts GPT4-P4 and P6, ChatGPT's accuracy could be further improved. Despite these promising results, it's important to note that the effectiveness of these prompts varies, which highlights the potential for further optimization and underlines the importance of testing on larger datasets.

Interestingly, GPT-P6N demonstrated a higher correlation with market returns than the actual sentiment, which hints at the model's superior understanding of overall market sentiment when processing all daily news at once. This could potentially be due to ChatGPT's ability to capture more nuanced sentiment variations across a set of news articles compared to individual human annotators who are assessing texts independently.

Moreover, models that generate sentiment scores ranging from -1 to 1 (N versions) showed a higher alignment with market movements. This suggests that these models are more adept at capturing subtle nuances in news sentiment, which might be more indicative of market trends. Conversely, GPT-P5, despite processing all daily news per forex pair, showed lower correlation with both market returns and true sentiment, thus emphasizing the importance of effective prompt engineering and rigorous evaluation of prompts before deployment.

The robust correlation observed between pairs GPT-P1 and P1N through to GPT-P6 and P6N further underscores the consistency and reliability of GPT's outputs, marking them as promising candidates for real-world applications.

\subsection{Potential Applications in Financial Services}

The findings of our study lend themselves to several practical applications in the financial services sector. With strategic prompt selection, the application of models like ChatGPT for financial sentiment analysis can offer actionable insights for predicting market trends. It is important to remember, though, that the choice of the ideal prompt could vary depending on the specific use case, the financial instrument under study, and the tolerance for prediction inaccuracies.

However, it's also crucial to clarify that the observed alignment between model sentiment scores and market movements does not necessarily imply predictive power over future price movements. Financial markets are complex systems influenced by a wide array of factors, including macroeconomic indicators, geopolitical events, and technical aspects. Recognizing this, a similar methodology to the one presented in this work could be applied with a more comprehensive set of input data. Instead of feeding the model with solely domain-specific news headlines, other relevant information such as economic indicators and additional market data could be included in the prompts, potentially yielding even richer insights. Therefore, while sentiment analysis using language models like ChatGPT can contribute valuable perspectives, they should be integrated into a more holistic approach to financial market analysis for optimum utility.

\subsection{Limitations and Future Work}

While our study reveals encouraging results, it also highlights some of the limitations of ChatGPT and suggests directions for future research. For instance, despite demonstrating substantial correlation with market sentiment, the models didn't entirely align with the market movements, indicating that sentiment only explains a part of the variations in market prices. This observation is an important reminder of the multifaceted nature of financial markets and points towards potential avenues for future research to further enhance these models.

As the landscape of LLMs continues to evolve with remarkable speed, it is imperative that future research delves into the comparison and assessment of the unique capabilities of both existing and emerging models. For instance, newer models such as GPT-4 \citep{openai2023gpt4} hold potential to enhance the performance exhibited by ChatGPT in our study. Furthermore, with the availability of open-source LLMs like BLOOM, a thorough evaluation of these models, their suitability for financial sentiment analysis, and their potential integration with commercial systems could prove to be a promising avenue of research. This continuous exploration of advancements in LLMs is necessary to fully leverage their potential in the dynamic and complex field of financial services.

In terms of processing time, further extensive testing under varied conditions is necessary to establish more accurate and consistent measures of response time. For operational applications that utilize LLMs services, such as the ChatGPT API, it would be advisable to have mechanisms in place to accommodate potential variations in response time due to factors like fluctuations in system load.

Taken together, the findings of our study underline the significant potential of ChatGPT in financial sentiment analysis. By pursuing the areas of future work identified here, we can move towards creating more sophisticated and accurate sentiment analysis models for financial markets.

\section{Conclusion}
\label{sec:6}

This work has explored the potential and adaptability of LLMs, specifically ChatGPT 3.5, in the challenging realm of financial sentiment analysis, focusing on the forex market. We employed a zero-shot prompting approach to evaluate the model's capacity to interpret and generate meaningful sentiment scores from financial text, bypassing the need for domain-specific fine-tuning. Our thorough examination was grounded in a meticulously curated and annotated dataset of forex-related news headlines, which we have made publicly available, providing a valuable resource to spur further research and innovation in the field.

Our evaluation, which extends beyond conventional metrics, revealed that prompts that accurately frame the intended task (short-term sentiment classification) show significant potential in understanding market behavior. This approach has demonstrated performance that exceeds that of FinBERT, which is an established model in the domain. Furthermore, our study underscored the critical importance of strategic prompt selection to enhance the performance and efficiency of sentiment analysis tasks, with ChatGPT demonstrating robustness and consistency across prompts.

However, while these results provide a promising outlook for the application of ChatGPT in financial sentiment analysis, they also suggest areas for further investigation. For instance, the superior understanding of overall market sentiment exhibited by prompts processing all daily news at once opens a pathway for future research exploring the integration of additional types of relevant financial data within the prompt.

In the broader landscape of AI tools in finance, our findings further substantiate the relevance and potential of LLMs. With the steady progress in this field, we anticipate more refined and robust models such as the GPT4, which could further enhance the performance of financial sentiment analysis.

While our study sheds light on the immense potential of LLMs in financial sentiment analysis, the path to real-world application is not without its challenges. Future work should look into addressing the varying time performance under different load conditions, establishing more accurate and consistent model performances.

In conclusion, our work contributes to a burgeoning field, offering a comprehensive evaluation of ChatGPT's application in financial sentiment analysis, and demonstrating its potential as a valuable tool in the realm of finance. We hope that our findings and the released dataset will serve as a springboard for further advancements in this domain. The exploration and refinement of LLMs in financial services is still in its early stages, and there is much more terrain to uncover.

\section*{CRediT authorship contribution statement}
\textbf{Georgios Fatouros:} Conceptualization, Data curation, Formal analysis, Investigation, Methodology, Software, Validation, Visualization, Writing – original draft. \textbf{John Soldatos:} Conceptualization, Funding acquisition, Supervision, Writing – review \& editing. \textbf{Kalliopi Kouroumali:} Data curation. \textbf{Georgios Makridis:} Writing – review. \textbf{Dimosthenis Kyriazis:} Funding acquisition, Project administration, Resources.

\section*{Declaration of Competing Interest}
The authors declare that they have no known competing financial interests or personal relationships that could have appeared to influence the work reported in this paper.

\section*{Acknowledgment}
Part of the research leading to the results presented in this paper has received funding from the European Union’s funded Project FAME under grant agreement no 101092639.




\begin{thebibliography}{39}
\expandafter\ifx\csname natexlab\endcsname\relax\def\natexlab#1{#1}\fi
\providecommand{\url}[1]{\texttt{#1}}
\providecommand{\href}[2]{#2}
\providecommand{\path}[1]{#1}
\providecommand{\DOIprefix}{doi:}
\providecommand{\ArXivprefix}{arXiv:}
\providecommand{\URLprefix}{URL: }
\providecommand{\Pubmedprefix}{pmid:}
\providecommand{\doi}[1]{\href{http://dx.doi.org/#1}{\path{#1}}}
\providecommand{\Pubmed}[1]{\href{pmid:#1}{\path{#1}}}
\providecommand{\bibinfo}[2]{#2}
\ifx\xfnm\relax \def\xfnm[#1]{\unskip,\space#1}\fi
\bibitem[{AI(2023)}]{jasper23report}
\bibinfo{author}{AI, J.}, \bibinfo{year}{2023}.
\newblock \bibinfo{title}{The ai in business trend report}.
\newblock \URLprefix \url{https://www.jasper.ai/blog/ai-business-trend-report}.
  \bibinfo{note}{accessed:May 26, 2023}.
\bibitem[{Arner et~al.(2015)Arner, Barberis and Buckley}]{arner2015evolution}
\bibinfo{author}{Arner, D.W.}, \bibinfo{author}{Barberis, J.},
  \bibinfo{author}{Buckley, R.P.}, \bibinfo{year}{2015}.
\newblock \bibinfo{title}{The evolution of fintech: A new post-crisis
  paradigm}.
\newblock \bibinfo{journal}{Geo. J. Int'l L.} \bibinfo{volume}{47},
  \bibinfo{pages}{1271}.
\bibitem[{Baker and Wurgler(2007)}]{baker2007investor}
\bibinfo{author}{Baker, M.}, \bibinfo{author}{Wurgler, J.},
  \bibinfo{year}{2007}.
\newblock \bibinfo{title}{Investor sentiment in the stock market}.
\newblock \bibinfo{journal}{Journal of economic perspectives}
  \bibinfo{volume}{21}, \bibinfo{pages}{129--151}.
\bibitem[{Bing(2012)}]{bing2012sentiment}
\bibinfo{author}{Bing, L.}, \bibinfo{year}{2012}.
\newblock \bibinfo{title}{Sentiment analysis and opinion mining (synthesis
  lectures on human language technologies)}.
\newblock \bibinfo{journal}{University of Illinois: Chicago, IL, USA} .
\bibitem[{Blaskowitz and Herwartz(2011)}]{blaskowitz2011economic}
\bibinfo{author}{Blaskowitz, O.}, \bibinfo{author}{Herwartz, H.},
  \bibinfo{year}{2011}.
\newblock \bibinfo{title}{On economic evaluation of directional forecasts}.
\newblock \bibinfo{journal}{International journal of forecasting}
  \bibinfo{volume}{27}, \bibinfo{pages}{1058--1065}.
\bibitem[{Bloomberg(2023)}]{Bloomberg2023}
\bibinfo{author}{Bloomberg}, \bibinfo{year}{2023}.
\newblock \bibinfo{title}{Bloomberg media distribution}.
\newblock \URLprefix
  \url{https://www.bloomberg.com/distribution/products/news/}.
  \bibinfo{note}{accessed:May 26, 2023}.
\bibitem[{Bollen et~al.(2011)Bollen, Mao and Zeng}]{bollen2011twitter}
\bibinfo{author}{Bollen, J.}, \bibinfo{author}{Mao, H.}, \bibinfo{author}{Zeng,
  X.}, \bibinfo{year}{2011}.
\newblock \bibinfo{title}{Twitter mood predicts the stock market}.
\newblock \bibinfo{journal}{Journal of computational science}
  \bibinfo{volume}{2}, \bibinfo{pages}{1--8}.
\bibitem[{Brown et~al.(2020)Brown, Mann, Ryder, Subbiah, Kaplan, Dhariwal,
  Neelakantan, Shyam, Sastry, Askell et~al.}]{brown2020language}
\bibinfo{author}{Brown, T.}, \bibinfo{author}{Mann, B.},
  \bibinfo{author}{Ryder, N.}, \bibinfo{author}{Subbiah, M.},
  \bibinfo{author}{Kaplan, J.D.}, \bibinfo{author}{Dhariwal, P.},
  \bibinfo{author}{Neelakantan, A.}, \bibinfo{author}{Shyam, P.},
  \bibinfo{author}{Sastry, G.}, \bibinfo{author}{Askell, A.}, et~al.,
  \bibinfo{year}{2020}.
\newblock \bibinfo{title}{Language models are few-shot learners}.
\newblock \bibinfo{journal}{Advances in neural information processing systems}
  \bibinfo{volume}{33}, \bibinfo{pages}{1877--1901}.
\bibitem[{Chen et~al.(2018)Chen, Fengler, H{\"a}rdle and Liu}]{chen2018textual}
\bibinfo{author}{Chen, C.}, \bibinfo{author}{Fengler, M.R.},
  \bibinfo{author}{H{\"a}rdle, W.K.}, \bibinfo{author}{Liu, Y.},
  \bibinfo{year}{2018}.
\newblock \bibinfo{title}{Textual sentiment, option characteristics, and stock
  return predictability} .
\bibitem[{Chen et~al.(2014)Chen, De, Hu and Hwang}]{chen2014wisdom}
\bibinfo{author}{Chen, H.}, \bibinfo{author}{De, P.}, \bibinfo{author}{Hu,
  Y.J.}, \bibinfo{author}{Hwang, B.H.}, \bibinfo{year}{2014}.
\newblock \bibinfo{title}{Wisdom of crowds: The value of stock opinions
  transmitted through social media}.
\newblock \bibinfo{journal}{The Review of Financial Studies}
  \bibinfo{volume}{27}, \bibinfo{pages}{1367--1403}.
\bibitem[{Dakhel et~al.(2023)Dakhel, Majdinasab, Nikanjam, Khomh, Desmarais and
  Jiang}]{dakhel2023github}
\bibinfo{author}{Dakhel, A.M.}, \bibinfo{author}{Majdinasab, V.},
  \bibinfo{author}{Nikanjam, A.}, \bibinfo{author}{Khomh, F.},
  \bibinfo{author}{Desmarais, M.C.}, \bibinfo{author}{Jiang, Z.M.},
  \bibinfo{year}{2023}.
\newblock \bibinfo{title}{Github copilot ai pair programmer: Asset or
  liability?}
\newblock \bibinfo{journal}{Journal of Systems and Software} ,
  \bibinfo{pages}{111734}.
\bibitem[{Devlin et~al.(2018)Devlin, Chang, Lee and Toutanova}]{devlin2018bert}
\bibinfo{author}{Devlin, J.}, \bibinfo{author}{Chang, M.W.},
  \bibinfo{author}{Lee, K.}, \bibinfo{author}{Toutanova, K.},
  \bibinfo{year}{2018}.
\newblock \bibinfo{title}{Bert: Pre-training of deep bidirectional transformers
  for language understanding}.
\newblock \bibinfo{journal}{arXiv preprint arXiv:1810.04805} .
\bibitem[{Evans and Lyons(2005)}]{evans2005currency}
\bibinfo{author}{Evans, M.D.}, \bibinfo{author}{Lyons, R.K.},
  \bibinfo{year}{2005}.
\newblock \bibinfo{title}{Do currency markets absorb news quickly?}
\newblock \bibinfo{journal}{Journal of International Money and Finance}
  \bibinfo{volume}{24}, \bibinfo{pages}{197--217}.
\bibitem[{Fatouros and Kouroumali(2023)}]{fxdataset}
\bibinfo{author}{Fatouros, G.}, \bibinfo{author}{Kouroumali, K.},
  \bibinfo{year}{2023}.
\newblock \bibinfo{title}{Forex news annotated dataset for sentiment analysis}.
\newblock \URLprefix \url{https://doi.org/10.5281/zenodo.7976208},
  \DOIprefix\doi{https://doi.org/10.5281/zenodo.7976208}. \bibinfo{note}{[Data
  set]}.
\bibitem[{Fatouros et~al.(2023)Fatouros, Makridis, Kotios, Soldatos, Filippakis
  and Kyriazis}]{fatouros2023deepvar}
\bibinfo{author}{Fatouros, G.}, \bibinfo{author}{Makridis, G.},
  \bibinfo{author}{Kotios, D.}, \bibinfo{author}{Soldatos, J.},
  \bibinfo{author}{Filippakis, M.}, \bibinfo{author}{Kyriazis, D.},
  \bibinfo{year}{2023}.
\newblock \bibinfo{title}{Deepvar: a framework for portfolio risk assessment
  leveraging probabilistic deep neural networks}.
\newblock \bibinfo{journal}{Digital finance} \bibinfo{volume}{5},
  \bibinfo{pages}{29--56}.
\bibitem[{George and George(2023)}]{george2023review}
\bibinfo{author}{George, A.S.}, \bibinfo{author}{George, A.H.},
  \bibinfo{year}{2023}.
\newblock \bibinfo{title}{A review of chatgpt ai's impact on several business
  sectors}.
\newblock \bibinfo{journal}{Partners Universal International Innovation
  Journal} \bibinfo{volume}{1}, \bibinfo{pages}{9--23}.
\bibitem[{Hossin and Sulaiman(2015)}]{hossin2015review}
\bibinfo{author}{Hossin, M.}, \bibinfo{author}{Sulaiman, M.N.},
  \bibinfo{year}{2015}.
\newblock \bibinfo{title}{A review on evaluation metrics for data
  classification evaluations}.
\newblock \bibinfo{journal}{International journal of data mining \& knowledge
  management process} \bibinfo{volume}{5}, \bibinfo{pages}{1}.
\bibitem[{Howard and Ruder(2018)}]{howard2018universal}
\bibinfo{author}{Howard, J.}, \bibinfo{author}{Ruder, S.},
  \bibinfo{year}{2018}.
\newblock \bibinfo{title}{Universal language model fine-tuning for text
  classification}.
\newblock \bibinfo{journal}{arXiv preprint arXiv:1801.06146} .
\bibitem[{\sortas{Mordor Intelligence}Mordor Intelligence(2022)}]{mordor22}
\bibinfo{author}{\sortas{Mordor Intelligence}Mordor Intelligence},
  \bibinfo{year}{2022}.
\newblock \bibinfo{title}{Ai in fintech market - growth, trends, covid-19
  impact, and forecasts (2023 - 2028)}.
\newblock \bibinfo{note}{URL:
  https://www.mordorintelligence.com/industry-reports/ai-in-fintech-market.
  Accessed:April 20, 2023}.
\bibitem[{Keynes(1937)}]{keynes1937general}
\bibinfo{author}{Keynes, J.M.}, \bibinfo{year}{1937}.
\newblock \bibinfo{title}{The general theory of employment}.
\newblock \bibinfo{journal}{The quarterly journal of economics}
  \bibinfo{volume}{51}, \bibinfo{pages}{209--223}.
\bibitem[{Kotios et~al.(2022)Kotios, Makridis, Fatouros and
  Kyriazis}]{kotios2022deep}
\bibinfo{author}{Kotios, D.}, \bibinfo{author}{Makridis, G.},
  \bibinfo{author}{Fatouros, G.}, \bibinfo{author}{Kyriazis, D.},
  \bibinfo{year}{2022}.
\newblock \bibinfo{title}{Deep learning enhancing banking services: a hybrid
  transaction classification and cash flow prediction approach}.
\newblock \bibinfo{journal}{Journal of big Data} \bibinfo{volume}{9},
  \bibinfo{pages}{100}.
\bibitem[{Liu et~al.(2021)Liu, Huang, Huang, Li and Zhao}]{liu2021finbert}
\bibinfo{author}{Liu, Z.}, \bibinfo{author}{Huang, D.}, \bibinfo{author}{Huang,
  K.}, \bibinfo{author}{Li, Z.}, \bibinfo{author}{Zhao, J.},
  \bibinfo{year}{2021}.
\newblock \bibinfo{title}{Finbert: A pre-trained financial language
  representation model for financial text mining}, in:
  \bibinfo{booktitle}{Proceedings of the twenty-ninth international conference
  on international joint conferences on artificial intelligence}, pp.
  \bibinfo{pages}{4513--4519}.
\bibitem[{Loughran and McDonald(2011)}]{loughran2011liability}
\bibinfo{author}{Loughran, T.}, \bibinfo{author}{McDonald, B.},
  \bibinfo{year}{2011}.
\newblock \bibinfo{title}{When is a liability not a liability? textual
  analysis, dictionaries, and 10-ks}.
\newblock \bibinfo{journal}{The Journal of finance} \bibinfo{volume}{66},
  \bibinfo{pages}{35--65}.
\bibitem[{OpenAI(2023)}]{openai2023gpt4}
\bibinfo{author}{OpenAI}, \bibinfo{year}{2023}.
\newblock \bibinfo{title}{Gpt-4 technical report}.
\newblock \href{http://arxiv.org/abs/2303.08774}{{\tt arXiv:2303.08774}}.
\bibitem[{Peters et~al.(2018)Peters, Neumann, Iyyer, Gardner, Clark, Lee and
  Zettlemoyer}]{peters2018deep}
\bibinfo{author}{Peters, M.E.}, \bibinfo{author}{Neumann, M.},
  \bibinfo{author}{Iyyer, M.}, \bibinfo{author}{Gardner, M.},
  \bibinfo{author}{Clark, C.}, \bibinfo{author}{Lee, K.},
  \bibinfo{author}{Zettlemoyer, L.}, \bibinfo{year}{2018}.
\newblock \bibinfo{title}{Deep contextualized word representations}, in:
  \bibinfo{booktitle}{Proceedings of the 2018 Conference of the North
  {A}merican Chapter of the Association for Computational Linguistics: Human
  Language Technologies, Volume 1 (Long Papers)},
  \bibinfo{publisher}{Association for Computational Linguistics},
  \bibinfo{address}{New Orleans, Louisiana}. pp. \bibinfo{pages}{2227--2237}.
\newblock \URLprefix \url{https://aclanthology.org/N18-1202},
  \DOIprefix\doi{10.18653/v1/N18-1202}.
\bibitem[{Poria et~al.(2017)Poria, Cambria, Bajpai and
  Hussain}]{poria2017review}
\bibinfo{author}{Poria, S.}, \bibinfo{author}{Cambria, E.},
  \bibinfo{author}{Bajpai, R.}, \bibinfo{author}{Hussain, A.},
  \bibinfo{year}{2017}.
\newblock \bibinfo{title}{A review of affective computing: From unimodal
  analysis to multimodal fusion}.
\newblock \bibinfo{journal}{Information fusion} \bibinfo{volume}{37},
  \bibinfo{pages}{98--125}.
\bibitem[{Poria et~al.(2016)Poria, Cambria and Gelbukh}]{poria2016aspect}
\bibinfo{author}{Poria, S.}, \bibinfo{author}{Cambria, E.},
  \bibinfo{author}{Gelbukh, A.}, \bibinfo{year}{2016}.
\newblock \bibinfo{title}{Aspect extraction for opinion mining with a deep
  convolutional neural network}.
\newblock \bibinfo{journal}{Knowledge-Based Systems} \bibinfo{volume}{108},
  \bibinfo{pages}{42--49}.
\bibitem[{Radford et~al.(2018)Radford, Narasimhan, Salimans, Sutskever
  et~al.}]{radford2018improving}
\bibinfo{author}{Radford, A.}, \bibinfo{author}{Narasimhan, K.},
  \bibinfo{author}{Salimans, T.}, \bibinfo{author}{Sutskever, I.}, et~al.,
  \bibinfo{year}{2018}.
\newblock \bibinfo{title}{Improving language understanding by generative
  pre-training} .
\bibitem[{Radford et~al.(2019)Radford, Wu, Child, Luan, Amodei, Sutskever
  et~al.}]{radford2019language}
\bibinfo{author}{Radford, A.}, \bibinfo{author}{Wu, J.},
  \bibinfo{author}{Child, R.}, \bibinfo{author}{Luan, D.},
  \bibinfo{author}{Amodei, D.}, \bibinfo{author}{Sutskever, I.}, et~al.,
  \bibinfo{year}{2019}.
\newblock \bibinfo{title}{Language models are unsupervised multitask learners}.
\newblock \bibinfo{journal}{OpenAI blog} \bibinfo{volume}{1},
  \bibinfo{pages}{9}.
\bibitem[{Sallam(2023)}]{sallam2023chatgpt}
\bibinfo{author}{Sallam, M.}, \bibinfo{year}{2023}.
\newblock \bibinfo{title}{Chatgpt utility in healthcare education, research,
  and practice: systematic review on the promising perspectives and valid
  concerns}, in: \bibinfo{booktitle}{Healthcare}, \bibinfo{organization}{MDPI}.
  p. \bibinfo{pages}{887}.
\bibitem[{Scao et~al.(2022)Scao, Fan, Akiki, Pavlick, Ili{\'c}, Hesslow,
  Castagn{\'e}, Luccioni, Yvon, Gall{\'e} et~al.}]{scao2022bloom}
\bibinfo{author}{Scao, T.L.}, \bibinfo{author}{Fan, A.},
  \bibinfo{author}{Akiki, C.}, \bibinfo{author}{Pavlick, E.},
  \bibinfo{author}{Ili{\'c}, S.}, \bibinfo{author}{Hesslow, D.},
  \bibinfo{author}{Castagn{\'e}, R.}, \bibinfo{author}{Luccioni, A.S.},
  \bibinfo{author}{Yvon, F.}, \bibinfo{author}{Gall{\'e}, M.}, et~al.,
  \bibinfo{year}{2022}.
\newblock \bibinfo{title}{Bloom: A 176b-parameter open-access multilingual
  language model}.
\newblock \bibinfo{journal}{arXiv preprint arXiv:2211.05100} .
\bibitem[{Schumaker and Chen(2009)}]{schumaker2009textual}
\bibinfo{author}{Schumaker, R.P.}, \bibinfo{author}{Chen, H.},
  \bibinfo{year}{2009}.
\newblock \bibinfo{title}{Textual analysis of stock market prediction using
  breaking financial news: The azfin text system}.
\newblock \bibinfo{journal}{ACM Transactions on Information Systems (TOIS)}
  \bibinfo{volume}{27}, \bibinfo{pages}{1--19}.
\bibitem[{Siering et~al.(2018)Siering, Muntermann and
  Rajagopalan}]{siering2018explaining}
\bibinfo{author}{Siering, M.}, \bibinfo{author}{Muntermann, J.},
  \bibinfo{author}{Rajagopalan, B.}, \bibinfo{year}{2018}.
\newblock \bibinfo{title}{Explaining and predicting online review helpfulness:
  The role of content and reviewer-related signals}.
\newblock \bibinfo{journal}{Decision Support Systems} \bibinfo{volume}{108},
  \bibinfo{pages}{1--12}.
\bibitem[{Tetlock(2007)}]{tetlock2007giving}
\bibinfo{author}{Tetlock, P.C.}, \bibinfo{year}{2007}.
\newblock \bibinfo{title}{Giving content to investor sentiment: The role of
  media in the stock market}.
\newblock \bibinfo{journal}{The Journal of finance} \bibinfo{volume}{62},
  \bibinfo{pages}{1139--1168}.
\bibitem[{Thoppilan et~al.(2022)Thoppilan, De~Freitas, Hall, Shazeer,
  Kulshreshtha, Cheng, Jin, Bos, Baker, Du et~al.}]{thoppilan2022lamda}
\bibinfo{author}{Thoppilan, R.}, \bibinfo{author}{De~Freitas, D.},
  \bibinfo{author}{Hall, J.}, \bibinfo{author}{Shazeer, N.},
  \bibinfo{author}{Kulshreshtha, A.}, \bibinfo{author}{Cheng, H.T.},
  \bibinfo{author}{Jin, A.}, \bibinfo{author}{Bos, T.}, \bibinfo{author}{Baker,
  L.}, \bibinfo{author}{Du, Y.}, et~al., \bibinfo{year}{2022}.
\newblock \bibinfo{title}{Lamda: Language models for dialog applications}.
\newblock \bibinfo{journal}{arXiv preprint arXiv:2201.08239} .
\bibitem[{Wolf et~al.(2020)Wolf, Debut, Sanh, Chaumond, Delangue, Moi, Cistac,
  Rault, Louf, Funtowicz et~al.}]{wolf2020transformers}
\bibinfo{author}{Wolf, T.}, \bibinfo{author}{Debut, L.}, \bibinfo{author}{Sanh,
  V.}, \bibinfo{author}{Chaumond, J.}, \bibinfo{author}{Delangue, C.},
  \bibinfo{author}{Moi, A.}, \bibinfo{author}{Cistac, P.},
  \bibinfo{author}{Rault, T.}, \bibinfo{author}{Louf, R.},
  \bibinfo{author}{Funtowicz, M.}, et~al., \bibinfo{year}{2020}.
\newblock \bibinfo{title}{Transformers: State-of-the-art natural language
  processing}, in: \bibinfo{booktitle}{Proceedings of the 2020 conference on
  empirical methods in natural language processing: system demonstrations}, pp.
  \bibinfo{pages}{38--45}.
\bibitem[{Wu et~al.(2023)Wu, Irsoy, Lu, Dabravolski, Dredze, Gehrmann,
  Kambadur, Rosenberg and Mann}]{wu2023bloomberggpt}
\bibinfo{author}{Wu, S.}, \bibinfo{author}{Irsoy, O.}, \bibinfo{author}{Lu,
  S.}, \bibinfo{author}{Dabravolski, V.}, \bibinfo{author}{Dredze, M.},
  \bibinfo{author}{Gehrmann, S.}, \bibinfo{author}{Kambadur, P.},
  \bibinfo{author}{Rosenberg, D.}, \bibinfo{author}{Mann, G.},
  \bibinfo{year}{2023}.
\newblock \bibinfo{title}{Bloomberggpt: A large language model for finance}.
\newblock \bibinfo{journal}{arXiv preprint arXiv:2303.17564} .
\bibitem[{Yeshayahou(2021)}]{Similarweb}
\bibinfo{author}{Yeshayahou, K.}, \bibinfo{year}{2021}.
\newblock \bibinfo{title}{Israeli co similarweb files for nyse ipo}.
\newblock \bibinfo{note}{URL:
  https://en.globes.co.il/en/article-israeli-co-similarweb-files-for-nyse-ipo-1001367823.
  Accessed:May 11, 2023}.
\bibitem[{Yue et~al.(2023)Yue, Au, Au and Iu}]{yue2023democratizing}
\bibinfo{author}{Yue, T.}, \bibinfo{author}{Au, D.}, \bibinfo{author}{Au,
  C.C.}, \bibinfo{author}{Iu, K.Y.}, \bibinfo{year}{2023}.
\newblock \bibinfo{title}{Democratizing financial knowledge with chatgpt by
  openai: Unleashing the power of technology}.
\newblock \bibinfo{journal}{Available at SSRN 4346152} .

\end{thebibliography}

\end{document}